\pdfoutput=1
\documentclass[11pt]{article}

\usepackage{acl}
\usepackage{times}
\usepackage{latexsym}

\newcommand{\subscript}[2]{$#1 _ #2$}

\usepackage[T1]{fontenc}
\usepackage[utf8]{inputenc}
\usepackage[usestackEOL]{stackengine}

\usepackage{microtype}
\usepackage{multicol}
\usepackage{soul}
\usepackage{multirow}
\usepackage{makecell}
\usepackage[utf8]{inputenc}
\usepackage{microtype}
\usepackage{graphicx}
\usepackage{soul}
\usepackage{float}
\usepackage{comment}
\usepackage{adjustbox}
\usepackage{booktabs}
\usepackage{amsmath}
\usepackage{mathabx}
\usepackage[shortlabels]{enumitem}
\usepackage[normalem]{ulem}
\usepackage{mathtools}
\usepackage{xspace}
\usepackage{array}
\usepackage{framed}
\usepackage{subcaption}
\usepackage{xcolor}

\usepackage{multicol}
\usepackage[ruled,vlined]{algorithm2e}
\usepackage{caption}
\usepackage{color,soul}
\usepackage{microtype}
\usepackage{dblfloatfix}
\usepackage{xcolor}

\newcommand{\dataset}{\textsc{Natural Instructions}}

\providecommand{\hanna}[1]{
    {\protect\color{blue}{[Hanna: #1]}}
}

\newcommand{\changed}[1]{#1}

\newcommand{\changednew}[1]{#1}

\newcommand{\cha}[1]{#1}

\usepackage{booktabs,arydshln}

\definecolor{purple}{rgb}{0.5,0,1}
\definecolor{dcyan}{rgb}{0.2,0.6,0.5}
\definecolor{darkgreen}{rgb}{0,200,0}
\definecolor{light-gray}{gray}{0.95} % used in table
\definecolor{darkgreen}{RGB}{0,140,0}
\definecolor{darkred}{RGB}{200,0,0}
\definecolor{lightgreen}{RGB}{231,255,219}
\definecolor{lightred}{RGB}{252,231,234}
\definecolor{lightyellow}{RGB}{250,253,191}
\definecolor{DarkRed}{RGB}{130,25,0}

\newcommand{\greentext}[1]{\colorbox{lightgreen}{#1}\xspace}
\newcommand{\yellowtext}[1]{\colorbox{lightyellow}{#1}\xspace}

\makeatletter
\def\adl@drawiv#1#2#3{%
        \hskip.5\tabcolsep
        \xleaders#3{#2.5\@tempdimb #1{1}#2.5\@tempdimb}%
                #2\z@ plus1fil minus1fil\relax
        \hskip.5\tabcolsep}
\newcommand{\cdashlinelr}[1]{%
  \noalign{\vskip\aboverulesep
           \global\let\@dashdrawstore\adl@draw
           \global\let\adl@draw\adl@drawiv}
  \cdashline{#1}
  \noalign{\global\let\adl@draw\@dashdrawstore
           \vskip\belowrulesep}}
\makeatother

\title{Task-transformation and Instruction Amelioration to Improve Large Language Model Performance}
\title{Speaking GPT3's Language: You Need to Ask the Right Question!}
\title{Speaking GPT3's Language: Task-transformation to Improve Model Response to Instructions}
\title{Speaking GPT3's Language: Prompt-transformation to Improve Model Response to Instructions}
\title{Speaking GPT3's Language: \\ \emph{Prompt-Reframing} to Improve Model Response to Instructions}
\title{Speaking GPT3's Language: \\ Reframing Prompts to  Improve Language Model's Response to Instructions}
\title{Speaking GPT3's Language: \\ Improving Model Response via \emph{Prompt-Reframing}}
\title{Speaking GPT3's Language: \\ An Analysis of Model Responses via \emph{Prompt-Reframing}}
\title{Speaking GPT3's Language: \\ An Analysis of Model Sensitivity to Prompt Framing}
\title{Speaking GPT3's Language: \\ An Analysis of Model Sensitivity to Prompt Framing}
\title{Speaking GPT3's Language: \\ Reframing Prompts to  Improve Model Response to Instructions}
\title{Reframing Tasks Definitions to the Language of Models}
\title{Speaking GPT3's Language}
\title{\name{}: Speaking GPT3's Language}
\title{Reframing Tasks Definitions to GPT3's Language}
\title{Reframing Tasks to Improve Prompting in GPT3}
\title{\name{} to Improve Prompting in GPT3}
\title{\name{}: Speaking GPT3's Language}
\title{Speaking GPT3's Language: Reframing Prompts to \\ Improve Model Response to Instructional Prompts}
\title{Reframing Descriptive Prompts to GPT3's Language}
\title{Reframing Descriptive Prompts to GPT's Language}
\title{Reframing Instructional Prompts to GPT's Language}
\title{
\vspace*{-0.5in}
{{\small \hfill ACL 2022 Findings}\\
\vspace*{.25in}} 
Reframing Instructional Prompts to GPT$k$'s Language}

\author{
Swaroop Mishra$^{3}\,$ Daniel Khashabi$^{1}\,$ \textbf{Chitta Baral}$^{3}\,$ \textbf{Yejin Choi}$^{1,2}\,$  \textbf{Hannaneh Hajishirzi}$^{1,2}$ 
\\\\
 $^1$Allen Institute for AI \; $^2$University of Washington \; 
 $^3$Arizona State University 
}

\begin{document}
\maketitle

\begin{abstract}

% How can model designers turn task instructions into  effective prompts for language models?
What kinds of instructional prompts are easier to follow for Language Models (LMs)?
We study this question by conducting extensive empirical analysis that shed light on important features of successful instructional prompts. 
\cha{
Specifically, we study several classes of 
% We propose several 
{\it reframing} techniques for 
manual reformulation of prompts into more effective ones. 
% model designers to manually create more effective  prompts. 
}
Some examples include decomposing a complex task instruction into multiple simpler tasks or itemizing instructions into sequential steps.
% \sout{For example, a complex task can be decomposed into multiple simpler tasks (Fig.\ref{fig:fig1a}) or the steps towards solving it could be itemized.}
% =low-level patterns about the target task, decomposing and itemizing instructions, stating the task constraints, and providing specialized instructions (examples in Table~\ref{tab:reframing:examples:main:text}). 
Our experiments compare the zero-shot and few-shot performance of LMs prompted with reframed instructions on 12 NLP tasks across 6 categories. % (question generation, classification, etc.). 
Compared with original instructions, our reframed instructions lead to  significant improvements across LMs with different sizes. 
% ,  underscoring the cross-model generality of these guidelines. %\sout{For example, \emph{the same} reframed prompts can boost GPT3 and GPT2 zero-shot performance by  17\% and} \hanna{gpt2 zero shot performance} 
\changednew{For example, \emph{the same} reframed prompts boost few-shot performance of GPT3-series and GPT2-series by 12.5\% and 6.7\% respectively averaged over all tasks.} %Furthermore, the same reframed prompts show similar performance gains across different, lighter model architectures (e.g., GPT2), underscoring the cross-model generality of these guidelines. We hope these empirically-driven techniques will pave the way for more effective ways to prompt LMs in the future.
Furthermore,  reframed instructions reduce the number of examples required to prompt LMs in the few-shot setting.
%The performance gains are particularly important on large language models, such as GPT3 where tuning models or prompts on large datasets is not feasible. Furthermore, these gains are not limited to GPT3; the same reframed prompts show superior performance gains across different, lighter model architectures (e.g., GPT2), underscoring the cross-model generality of these guidelines. 
We hope these empirically-driven techniques will pave the way 
\cha{
towards more effective future prompting algorithms.
}
\end{abstract}

\section{Introduction}
\begin{figure}
    \centering
    \includegraphics[scale=0.65, trim=1.5cm 1cm 1cm 1cm]{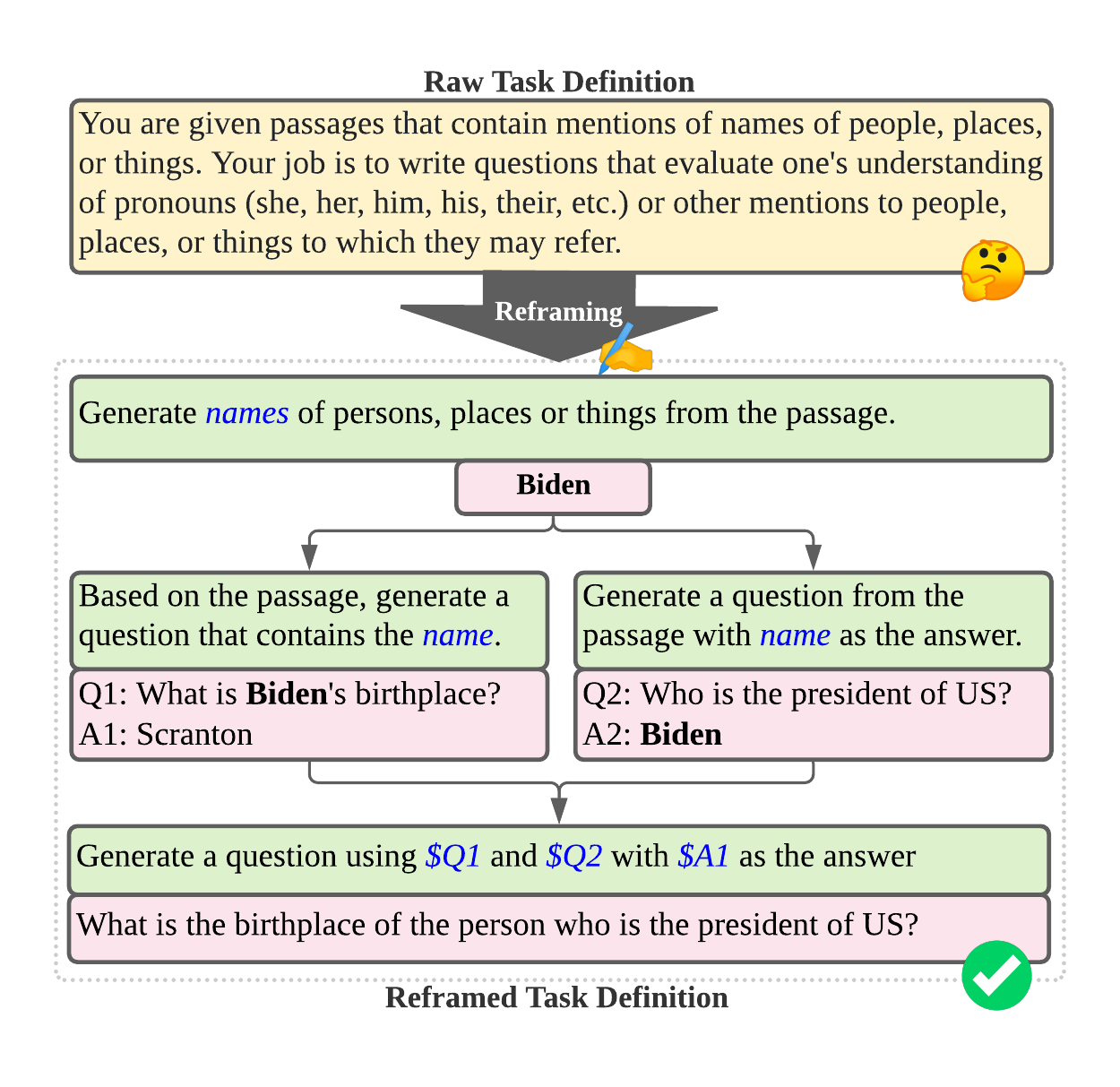}
            \caption{GPT3 has difficulty in writing questions that require entity coreference resolutions based on a single lengthy prompt (top, in \yellowtext{yellow}),
        however, it succeeds in solving a manually \emph{reframed} task that has four simpler sub-steps (bottom, in \greentext{green}).}
     \label{fig:fig1a}
\end{figure}

Prompting language models (LMs) \cite{liu2021promptingsurvey} has made NLP modules accessible to  non-expert users through plain text \cha{instructions\footnote{
  We focus on instructional prompts~\cite{efrat2020turking} as opposed to exemplar prompts which are already well-studied~\cite{brown2020gpt3,lu2021fantastically}. 
} of NLP tasks. 
Such task instructions written by non-expert users are often long and contain abstract descriptions which are not easy to follow for LMs, as evident by their low performance ~\cite{efrat2020turking,mishra2021natural}. 
However, it is not quite clear whether this is due to the inherent difficulty of the target tasks or an artifact of the complex phrasing of their language instructions. 
}

\cha{
In this analysis, we aim to  understand the sensitivity of LMs to the framing of instructional prompts. 
% In this work, 
% we analyze the sensitivity of LMs the framing of task instructions they are prompted with.
% the challenges that give rise to poor performance of LMs when prompted with complex task instructions
% \changednew{(Table \ref{tab_error_analysis_main_text})} 
% \hanna{I replace the next sentence with the following: In particular, we study several {\it reframing} techniques to frame instructional prompts differently so that LMs achieve better understanding of the task. } 
% \sout{We study this objective via several class of  \emph{reframing} instructional prompts.}
In particular, we study several {\it reframing} techniques to frame instructional prompts differently so that LMs achieve better understanding of the task.
% and provide  guidelines to manually \emph{reframe} them to effectively prompt LMs. 
These reframing techniques are motivated by various empirical intuitions such as ease of understanding concise and concrete instructions and those that contain} little abstract statements about human commonsense or their background knowledge. 
For example, Fig.\ref{fig:fig1a} shows a reframing example which involves decomposing a task into multiple sub-tasks. 
The intended task here is
writing questions that require entity coreference~\cite{dasigi2019quoref}.
While GPT3 fails in solving the original task instruction 
(the yellow box at the top), it succeeds when the task is decomposed to four simpler and easier sub-tasks.

\cha{
% Our analysis on reframing consists of five techniques
% We propose five types of reframing techniques that 
% can 
% be applied manually by model designers. 
% that we apply manually. 
We provide analysis for five diverse reframing techniques. 
}
These include incorporating low-level patterns about the target task, decomposing and itemizing instructions, stating the task constraints, and providing specialized instructions (examples in Table~\ref{tab:reframing:examples:main:text}).

\begin{figure}
    \centering
    \includegraphics[scale=0.5, trim=0.7cm 0cm 0cm 0cm]{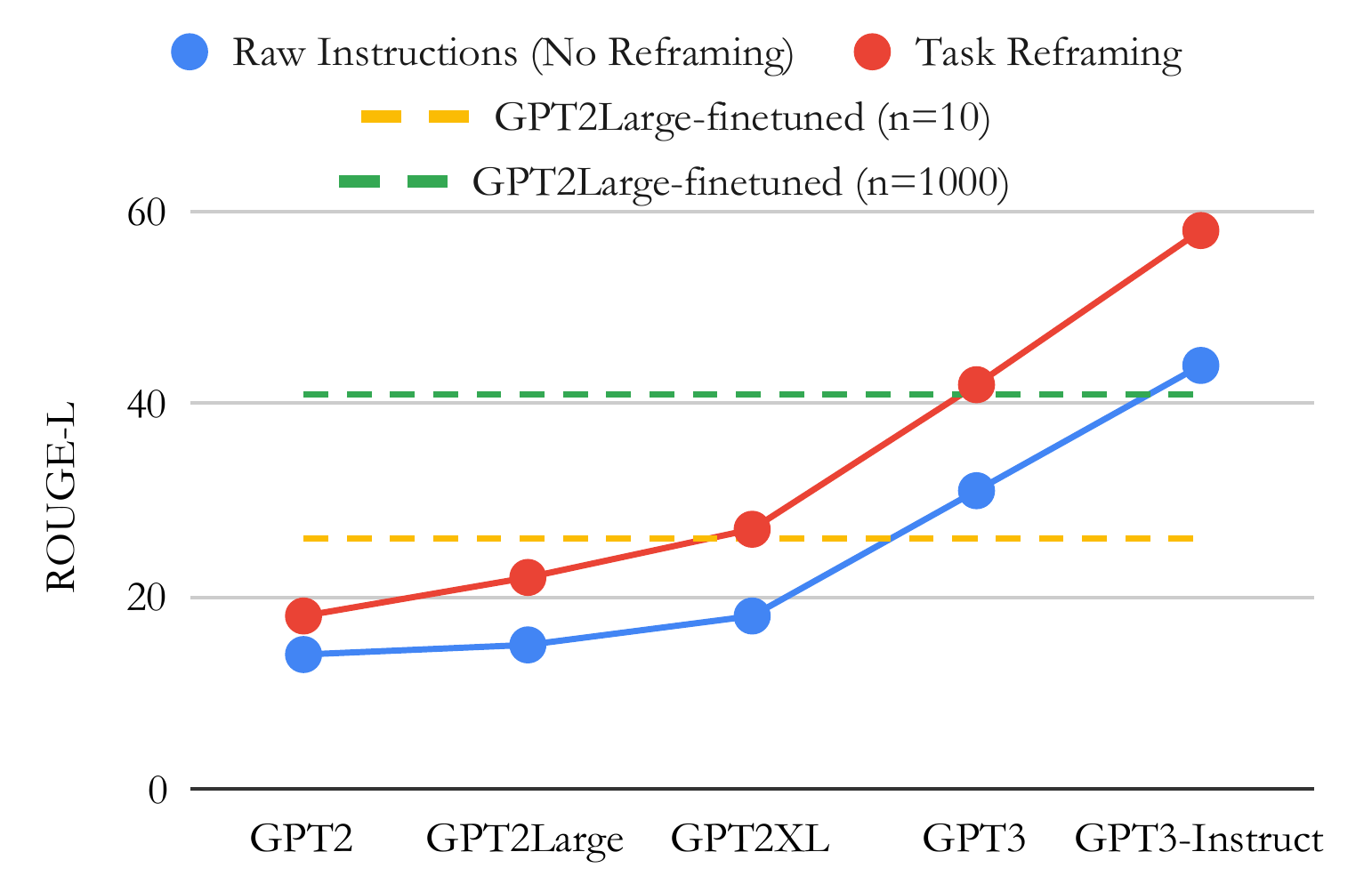}
        \caption{
        Across a variety of model sizes, {\color{red}reframed prompts} consistently show considerable performance gain over {\color{blue} raw task instructions (no reframing)} \changed{in a few-shot learning setup}. 
        % Cross-model generalization across several models in the GPT-Series. 
        % \name{} (red line) results in considerable performance improvement over original task instructions i.e. without reframing (blue line) across all models. 
        Since fine-tuning GPT3 is prohibitively expensive, we show the performance of fine-tuning smaller models ({\color{darkgreen}horizontal} {\color{orange} lines}). 
        % For comparison, we show the performance of fine-tuning , 
        This results indicates that
        \changed{
            % for a fixed compute budget, 
            \emph{evaluating} reframed prompts on a large model like GPT3-instruct (red line) might be more effective that \emph{fine-tuning}} a smaller model like GPT2Large (green line) with 200$\times$ more data.
            % This results indicates that it might be more effective for a model designer to reframe a task prompt for a large model like GPT3-instruct (red line) than performing fine-tuning for a smaller model like GPT2Large (green line) with 200$\times$ more data.
            Details of the experiments in \S\ref{sc:experiments}. 
        }
        \label{fig:fig1b}
\end{figure}

\cha{
We analyze reframed instructions over 
% We evaluate reframing techniques on over 
}
12 tasks from \dataset{}~\cite{mishra2021natural}, which contains a variety of NLP tasks and their instructions. 
Empirically, we compare the quality of LMs (GPT2/3~\citealt{radford2019language,brown2020gpt3}) in two settings: raw vs reframed instructions.
In particular, 
we observe that the reframed prompts have notable performance gains over raw instructions (the gap between the red and blue trends in Fig.\ref{fig:fig1b}) with an  average of 14\%  and 17\% gains when using GPT3-instruct in the few-shot and zero-shot setups, respectively.
% \changed{
% This gap is even larger in zero-shot setup where no additional examples are supplied as part of the input prompt (Table~\ref{tab:main}). 
% }
Furthermore, the average gains across tasks remain consistent across different models hinting at consistency of reframed prompts on various architectures. 
This is in contrast to the \changed{widely-used} fine-tuning approaches which need to be performed separately for each model.
Reframing prompts by model designers can be  particularly  effective  when evaluated on \underline{large} LMs, 
% where fine-tuning is
where fine-tuning can be
prohibitively expensive (such as GPT3). 
In particular, we observe that, reframed prompts on GPT3-instruct score roughly $17\%$  higher  than GPT2Large \changed{that is} supervised with $1k$ instances (i.e., 200$\times$ more data).

% and a few examples~\cite{liu2021promptingsurvey}.
% In particular, engineering instructional prompts
% have been studied in the context of tasks such as classification~\cite{jiang2020whatmodelsknow,schick2021exploiting}, where instructions  are often limited to short and simple phrasings. 
% has produced successful results in several NLP tasks with short and concrete descriptions, such as classification~\cite{jiang2020whatmodelsknow,schick2021exploiting}.
% Compared to the recent literature, our reframing techniques broaden the scope of existing prompt engineering approaches~\cite{petroni2019language,schick2021exploiting} which generally focus on surface-level changes to the original prompt. 

% \cha{
% There are dozens of works use \emph{manually} engineered short prompts~\cite{petroni2019language,jiang2020whatmodelsknow,schick2021exploiting} for tasks such as classification or text in-filling. However, there has not been much analysis on effective framing of instructional prompts -- which is the focus of this work.
% }

\cha{
While reframing instructions are not algorithmic, nonetheless, we view this systemic analysis as a preliminary stepping stone in this direction. 
% which, which we hope will guide the future progress toward algorithmic solutions.
We hope that this study will lead to the development of algorithmic better few-shot learning methods that generalize across models, thereby leading to more effective ways of reaping the investments already poured into creating massive LMs. 
% Using our guidelines \changed{for reframing}, model designers can come up with new reframed tasks in a matter of minutes. 
% We view our work (systemic analysis of effective language prompts) also as a stepping stone in this direction which, we believe, will guide the future progress toward algorithmic solutions.
}

\noindent \textbf{Contributions:} 
(a) \changed{This work is inspired by the sensitivity of LMs to the framing of their instructional prompts.}
Driven by many empirical analysis, we identify several guidelines for model designers to \emph{reframe} instructional prompts and provide illustrative use cases associated with each type of reframing technique. 
(b) Extensive experiments on diverse tasks show that reframing gives rise to superior 
% few-shot learning 
performance \changed{and improved sample complexity} over raw task instructions, across a range of models sizes. 
% and shows evidence for cross-model generalization. 
% Additionally, we find evidence that reframing prompts on large models might outweigh fine-tuning smaller models.  
% \daniel{empirical: over raw instructoons and over smaller fine-tuned models. }
(c) Our experiments  quantify the contribution of the prompting techniques and analyze various parameters that contribute to their success.

%An alternative scenario might be fine-tune \underline{smaller} models that are within computation budget, such as, GPT2Large fine-tuned on examples of our target tasks (Fig.~\ref{fig:fig1b}; vertical lines). 

% which are below the performance of reframed prompts on GPT3. 
%In sum, the trends show that if one is faced with two options of (a) fine-tuning small model (e.g., GPT2Large) vs (b) reframing prompts on a large model (e.g., GPT3), the latter can be a more effective approach. 

\section{Related Work}
% also talk about noah smith paper on shortformer and transformer limitation.
Our work is related to designing discrete prompts and tuning continuous prompts in recent literature. 

\vspace{.1cm} \noindent \textbf{Discrete Prompts}
Constructing effective discrete prompts for language models to perform NLP tasks is an active area of research~\cite{schick2020few, le2021many, tam2021improving, logan2021cutting, reynolds2021prompt}. 
\cha{
Most such works focus on light-weight changes 
% change surface-level changes 
to the original prompt~\cite{liu2021promptingsurvey}. 
Unlike the earlier literature, 
we focus on framings of complex instructions, 
which often lead to reframed prompts that are often very different from the original raw instructions.  
% At a high level, our reframing techniques extend prompt engineering approaches ~\cite{liu2021promptingsurvey} to enable language models to understand and follow complex task instructions, and fill in several gaps in the prompt engineering literature in the following ways: 
% (1) Most recent prompt-engineering approaches make strong assumptions about their target tasks which make them non-trivial to generalize to any NLP task. 
% For example, they are studied in the context of classification~\cite{petroni2019language,schick2021exploiting} or  applied as prompt composition and prompt decomposition for relation extraction~\cite{han2021ptr} and named entity recognition~\cite{cui2021template} tasks, respectively. 
While our proposed prompt-reframing is not quite algorithmic, the principles behind them are relatively simple, which can hopefully motivate algorithmic solutions in future. 
% and enable model designers to reframe a variety of tasks beyond classification. 
%This way, reframing broadens the scope of prompt-engineering.
}
% Existing proposals to discover prompts~\cite{shin2020eliciting, jiang2020whatmodelsknow} lead to results that are model-specific, i.e., the resulting prompts are not shown to generalize across models. 
% We show evidence of cross-model generalization for the reframed prompts. 

\cha{Our goal is fundamentally different from the meta-training with instructions~\cite{mishra2021natural,sanh2021multitask, wei2021finetuned}. 
% which 
% utilize assume labeled data (discrete prompts of thousands of tasks) for 
% fine-tune better instruction-following models. 
% a model for better generalization. 
% This is categorically different from the purpose of our analysis work -- given a model with frozen weights, we carefully analyze what types of language prompts tend to be more effective. 
Such approaches depend on labeled data (language prompts for thousands of tasks) which can be costly to collect. 
Additionally, they require
% as well as the ability to fine-tune models 
fine-tuning models which can be costly for larger LMs. 
% This is an important distinction since with models consistently getting larger, the concerns regarding the   carbon footprint of fine-tuning is a legitimate reason to look for other alternatives~\cite{schwartz2020green}. 
% We believe the community will benefit from considering all these complementary approaches towards more effective prompting of LMs. 
Exploring effective framings of language instructions can provide alternative ways of utilizing LMs.
}

\vspace{.1cm} \noindent \textbf{Continuous Prompts}
Tuning continuous prompts leads to the making of space-efficient models compared to fine-tuning model parameters \cite{liu2021gpt, lester2021power}. 
% reynolds2021prompt, li2021prefix, ben2021pada 
Despite being algorithmic, these models require propagating gradient information across the whole architecture, leading to high computational costs, which is a key bottleneck when it comes to large LMs such as GPT3. 
While our proposal 
% is not quite algorithmic and 
requires human intervention, it provides model designers with several relatively easy rules-of-thumb to come up with language prompts that work effectively with large LMs.

\begin{table*}
    \small
    \centering
   %  \resizebox*{!}{0.98\textheight}{%
    \begin{tabular}{p{0.42cm}p{14cm}}
    \toprule
         & \textit{Raw task definitions and their reframed counterpart} \\ 
        \midrule
        \parbox[t]{2mm}{\multirow{3}{*}{\rotatebox[origin=c]{90}{
         \textsc{\textcolor{purple}{\parbox{3.7cm}{\small \centering Pattern \\ Reframing}}}
         }}}
        & \textcolor{brown}{\textit{\textbf{Raw Task:}}}  \textit{Craft a question which requires commonsense to be answered. Based on the given context, craft a common-sense question, especially those that are LONG, INTERESTING, and COMPLEX. The goal is to write questions that are easy for humans and hard for AI machines! To create such questions, here are some suggestions: A. What may (or may not) be the plausible reason for an event? B. What may (or may not) happen before (or after, or during) an event? C. What may (or may not) be a plausible fact about someone (or something)? D. What may (or may not) happen if an event happens (or did not happen)? You can also create other types of questions.} \newline \textbf{Input}: Context:<> \quad \textbf{Expected Output}: Question:<>\\ [0.2cm]
        \cdashlinelr{2-2}
        \rule{0pt}{0.4cm}  & \textcolor{brown}{\textit{\textbf{Reframed Task:}}}
        \textcolor{blue}{\textit{Use 'what may happen', 'will ...?', 'why might', 'what may have caused', 'what may be true about', 'what is probably true about', 'what must' and similar phrases in your question based on the input context.}}\newline \textbf{Input}: Context:<> \quad \textbf{Expected Output}: Question:<>\\
        
        \midrule
        {\multirow{3}{*}{\rotatebox[origin=r]{90}{
         \textsc{\textcolor{purple}{\parbox{2.2cm}{\small \centering Itemizing \\ Reframing}}} 
         }}} &\textcolor{brown}{\textit{\textbf{Raw Task:}}} {\textit{Follow the instructions to produce output with the given context word. Do <>. Do <>. Don’t <>}} \newline \textbf{Input}: Context word <> \quad \textbf{Expected Output}: Long text <> \\ [0.2cm]
         \cdashlinelr{2-2}
        \rule{0pt}{0.4cm}  & \textcolor{brown}{\textit{\textbf{Reframed Task:}}} {\textit{Follow instructions below to produce output based on the given context word.}} \newline - \textcolor{blue}{Do <> \newline - Do <> \newline - Do <>}\newline \textbf{Input}: Context word <> \textbf{Expected Output}: Long text <>\\
         
         \midrule
        {\multirow{3}{*}{\rotatebox[origin=c]{90}{
         \textsc{\textcolor{purple}{\parbox{7.1cm}{\small \centering Decomposition\\ Reframing}}}
         }}} 
         & \textcolor{brown}{\textit{\textbf{Raw Task:}}} \textit{In this task, based on the given context word, you need to create a pair of sentences each containing a blank (\_) and their corresponding answer. The sentence pair should look similar, and should be about two related but different objects; for example "trophy" and "suitcase". Also, the sentences must be different in terms of trigger words (e.g., "small" and "big") which express contrasting properties about the two objects.} \newline \textbf{Input}: Context word:<> \quad \textbf{Expected Output}: Question 1: <> Answer 1: <> Question 2: <> Answer 2: <> \\ [0.2cm]
         \cdashlinelr{2-2}
         \rule{0pt}{0.4cm} & \textcolor{brown}{\textit{\textbf{Reframed Task:}}} 
         \vspace{0.2cm}
         \newline \textbf{Subtask 1.} \textit{ \textcolor{blue}{ Write 2 objects based on the given context word.}} \newline \textbf{Input}: Context word:<> \quad \textbf{Expected Output}: \textcolor{red}{Objects: <>} 
         \vspace{0.2cm}
         \newline \textbf{Subtask 2.} \textit{ \textcolor{blue}{ Write a sentence by connecting objects with a verb.}} 
         \newline \textbf{Input}: \textcolor{red}{Objects: <> } \quad  \textbf{Expected Output}: \textcolor{red}{Sentence: <>} 
         \vspace{0.2cm}
         \newline  \textbf{Subtask 3.} \textcolor{blue}{\textit{Create a fill in the blank question from the sentence where object 1 will fit the blank.}} 
         \newline \textbf{Input}: \textcolor{red}{Object 1: <>,Sentence: <>} \quad \textbf{Expected Output}: \textcolor{red}{Question: <>} 
         \vspace{0.2cm}
         \newline \textbf{Subtask 4.} \textcolor{blue}{\textit{ Change the given question so that answer flips to object 2 in the question.}} 
         \newline \textbf{Input}: \textcolor{red}{Object 2: <>, Sentence: <>, Question: <>} \quad  \textbf{Expected Output}: \textcolor{red}{Question: <>}
         \vspace{0.2cm}
         \newline \textbf{Subtask 5.} \textcolor{blue}{\textit{ Generate both questions and answers:}} 
         \newline \textbf{Input}: \textcolor{red}{Question 1: <> Object 1: <> Question 2: <> Object 2: <>} 
         \newline \textbf{Expected Output}: Question 1: <> Answer 1: <> Question 2: <> Answer 2: <>\\
        
        \midrule
         {\multirow{3}{*}{\rotatebox[origin=l]{90}{
         \textsc{\textcolor{purple}{\parbox{2.3cm}{\small \centering Restraining\\Reframing}}} 
         }}} & \textcolor{brown}{\textit{\textbf{Raw Task:}}}... \textit{What is the type of the answer corresponding to the given question? Number, Date, or Span?}...\newline \textbf{Input}: Passage: <>. Question: <> \quad \textbf{Expected Output}: <Number/Date/Span> ...\\ [0.2cm] 
         \cdashlinelr{2-2}
         \rule{0pt}{0.4cm}  & \textcolor{brown}{\textit{\textbf{Reframed Task:}}}... \textit{What is the type of the answer corresponding to the given question? Number, Date, or Span?}... \newline \textbf{Input}: Passage: <> Question: <> \textcolor{blue}{\textit{Answer either Number, Date or Span?}} \textbf{Expected Output}:<Number/Date/Span>\\
        
        \midrule
        {\multirow{3}{*}{\rotatebox[origin=c]{90}{
         \textsc{\textcolor{purple}{\parbox{2.1cm}{\small \centering Specialization Reframing}}} 
         }}} & \textcolor{brown}{\textit{\textbf{Raw Task:}}} \textit{Answer the following question ... <Not so important Text> ...} \newline \textbf{Input}: Question <> \quad \textbf{Expected Output}: Answer <>  \\ [0.2cm]
         \cdashlinelr{2-2}
         \rule{0pt}{0.4cm} & \textcolor{brown}{\textit{\textbf{Reframed Task:}}}\textcolor{blue}{\textit{Calculate answer to the following question.} \textit{You need to either add or subtract numbers associated with two objects present in the question.}} \newline \textbf{Input}: Question <> \quad \textbf{Expected Output}: Answer <> \\
        \bottomrule
\end{tabular}
% }
\caption{Examples of various reframing techniques. \textit{Italicized} text represents the prompt. Change in prompt and  example in the transformed task are indicated with \textcolor{blue}{blue} and \textcolor{red}{red} markings, respectively. 
}
    \label{tab:reframing:examples:main:text}
\end{table*}

\section{Prompt Reframing}
This section describes our reframing principles and then describes the guidelines to  operationalize them. Reframing principles are obtained by probing instructions of various tasks in the training split of \dataset{}~\cite{mishra2021natural} to understand different failure modes associated with prompting in GPT3. 
% \daniel{
% Hanna's suggestion: we need to justify why NI (and not any other existing dataset) -- need to discuss the complexity and length of the instructions. 
% }
% and subsequently come up with reframing techniques based on our empirical observations using raw and reframed tasks.
% We propose \name{} to reframe the original task instructions \cite{mishra2021natural} that often contain all details to help models understand a task better. 
% In our experimentation, authors reframe tasks manually with a small development set of 10 samples before applying to the test set of a task.

\paragraph{Motivation from GPT3's Failures} 
We observe that GPT3 fails to follow instructions when it is provided with long prompts that often contain repeated information, abstract notions, analogies, complex statements  requiring human commonsense and their domain knowledge (see examples in Table \ref{tab:reframing:examples:main:text} and \ref{tab_error_analysis_main_text}).
Humans typically find these helpful for describing their tasks. 
% All these information help humans in explaining a task: 
For example, some content intended to motivate the task or repetition for the sake of emphasis, might be unnecessary or even redundant  for a model.
% repeated information is to emphasize, analogy is to relate to various known concepts, complex instruction with terms associated with background knowledge is to summarize and express information in a human-friendly language. However, machines do not benefit from these information unlike humans.

\subsection{Reframing Principles}
We observe that short prompts that contain concrete statements and avoid terms associated with background knowledge improve GPT3's response to instructions. We recursively apply this observation and provide a set of reframing principles to resolve various issues on GPT3's failures with prompting, backed by extensive empirical analysis on GPT3.\footnote{ The principles have light resemblance to how basic tasks are formulated and taught to kids.}

\begin{enumerate}[noitemsep, label=(\subscript{C}{{\arabic*}}),leftmargin=25pt]
    
    \item \label{cp2} \emph{Use Low-level Patterns}: Instead of using terms that require background knowledge to understand, use various patterns about the expected output. 
    % possible output types as low-level patterns in instruction.
    
    \item \label{cp4} \emph{Itemizing Instructions}: Turn descriptive attributes into bulleted lists. If there are any negation statements, turn them into assertion statements. 
    
    \item \label{cp1} \emph{Break it Down}: Break down a task into multiple simpler tasks, wherever possible. 
    % Continue breaking down until a task follows all the above design principles.
    
    \item \label{cp3} \emph{Enforce Constraint}: Add explicit textual statements of output constraints. 
    % model to answer to the point and in the intended format.
    
    \item \label{cp5} \emph{Specialize the Instruction}: Customize the instructions so that they directly speak to the intended output.

    % to the task instead of having a  for every task such as "answer the question". For example, a numerical reasoning task can have the instruction "calculate answer". 
    
    % \item \label{cp6} \emph{Add Diverse Examples if Instruction is Inadequate}: Avoid examples if plain instruction can do the task. Examples can produce high variance in model prediction \cite{zhao2021calibrate, lu2021fantastically}. If a task is difficult to describe in instructions without using domain-specific words, use examples but try to have diverse example types (in terms of question type or answer type) as much as possible.
\end{enumerate}

We operationalize each of the above principles in terms of 5 \emph{reframing} techniques. 
% In total, we have 5 reframing techniques that collectively define \name{}. 
The degree of reframing (the amount of change applied to the raw instructions) varies significantly across the reframing techniques: the simplest one adds an enforcement statement at the end
% right before a model is asked to predict output 
whereas 
% reframing at 
the other extreme involves completely changing the task as a whole (e.g., decomposing it into multiple tasks). 
% or transforming it to an equivalent/ analogous task. 

\subsection{Reframing Techniques}
\label{subsec: reframtech}

We explain each of the reframing techniques in three parts (1) \emph{model failure} states a potential weakness of LM with reference to examples in Table~\ref{tab_error_analysis_main_text} (2) \emph{approach} describes our suggested approach and intuition behind it, according to our empirical observations  
% observation: this explains our observation on how reframing tries to overcome the weakness and make the task easier, 
(3) \emph{example} illustrates the application of the suggested technique in reference to Table~\ref{tab:reframing:examples:main:text}.
In designing these techniques, we used a development set that contains all the positive examples included as part of the instructions of each task in \dataset{}.
%  all the positive examples included as part of the instructions of each task 
% associated with a task 
% in \dataset{} (an average of $\sim5$ examples per task). 
% On an average, each task contains $\sim5$ positive examples in \dataset{}, hence our dev set has $\sim$5 samples. 
% just change the GPT3 parameters such as engine, temperature and stop condition

\begin{comment}
\begin{table}[t]
    \small 
    \centering
    \begin{tabular}{ll}
        \toprule
        Reframing Type  & Applicability  \\
        \midrule
        \textsc{Pattern } & All types of tasks and instructions \\ 
        \textsc{Itemizing } &  Multi-step instructions \\
        \textsc{Decomposition } &  Multi-step instructions \\ 
        \textsc{Restraining }   & Classification and extractive QA \\ 
        \textsc{Specialization } & Special operations and reasoning \\ 
        \bottomrule
    \end{tabular}
    \caption{Applicability of each type of \emph{reframing} techniques. Special operations involve various operations involved in QA such as addition, subtraction of numbers. \hanna{where is reference to Table 2?}
    % in \name
    }
    \label{tab:taskcategories}
\end{table}
\end{comment}

\subsubsection{\textsc{Pattern Reframing}}
\textbf{Model failure} While humans have an incredible ability in understanding and acting with respect to abstract 
% and high-level 
descriptions, \changednew{LMs tend to ignore most of them or just repeat the content of such instructions in their output (\textit{copy instruction} in Table  \ref{tab_error_analysis_main_text}.)}\\ 
% part of the instructions.\\
% it is hard for models like GPT3 to understand and follow such instructions. \\
\textbf{Approach}
Find low-level patterns among the dev set examples and extrapolate those by adding similar patterns~\ref{cp2}. \\
% We observe that, this way of converting high-level instructions into low-level instructions that does not contain terms associated with domain-specific/ commonsense knowledge improves model performance \ref{cp2}.\\
\textbf{Example} Table~\ref{tab:reframing:examples:main:text} (row 1) illustrates the CosmosQA \cite{huang2019cosmos} question generation task. 
The raw task instruction consists of various high-level statements such as ``commonsense'', ``complex'', ``interesting'', ``easy for humans and hard for AI machines'', whereas the reframed task consists of various low-level patterns about the expected output such as ``what may happen'', ``in the future, will..'', ``why might'', which generally 
% and asks model to generate similar phrases; this 
improve GPT3's performance in generating valid questions.
%instead of using complex concept, we enumerate a concept as a set of patterns
% do some largest, smallest task

\subsubsection{\textsc{Itemizing Reframing}}
\label{subsubsec:strref}
\textbf{Model failure} LMs cannot follow long paragraphs stating multiple requirements \changednew{(\textit{first instruction bias} in Table \ref{tab_error_analysis_main_text})} and 
do not perform well when the requirements are formulated as a negative statement (\changednew{\textit{negation challenge} in Table \ref{tab_error_analysis_main_text})}.
% have difficulty often fail to follow negative instructions. 
\\
\textbf{Approach} 
Turn long descriptions into bulleted lists of several statements~\ref{cp4}. 
% In case of multistep instructions, we observe that providing instructions in a structured format in the form of a list instead of a paragraph helps models \ref{cp4}.
Additionally, turn negative statements to positive ones. 
% improves model performance. 
For example, reformulate ``don't create questions which are not answerable from the paragraph'' into ``create questions which are answerable from the paragraph''. \\
% We observe that converting negative instructions to positive instructions improves model performance. For example: instead of saying ``don't create questions which are not answerable from the paragraph" we can say ``create questions which are answerable from the paragraph". \\
\textbf{Example} Table \ref{tab:reframing:examples:main:text} (row 2) illustrates the WinoGrande~\cite{sakaguchi2020winogrande} sample generation task where the raw instructions contain several requisites (do's and don'ts) that are hard for models to follow.
% some Do's and Don'ts. 
% Model typically fails to follow these instructions, especially the later ones and Don'ts. 
Reframing the instructions into 
% replacing Don'ts by Dos, and presenting in 
a structured list improves the model response. % \paragraph{Analogy reframing}
% % ModularQA, this is not pure example based. If not enough, create a synthetic task based on the analogy example.

\subsubsection{\textsc{Decomposition Reframing}}
\textbf{Model failure} Tasks with implicit multi-step reasoning are challenging for models, even after itemizing reframing  (\ref{subsubsec:strref}) (\changednew{\textit{multi-step task challenge} in Table \ref{tab_error_analysis_main_text}}).\\
\textbf{Approach} Wherever possible, decompose a task into multiple different sub-tasks which can be executed either sequentially or in parallel \ref{cp1} and hence, make them 
% . Our intention behind this decomposition is to make the sub-tasks 
relatively easier for models. \\
% We observe that decomposition is highly effective in improving response of models to instructions.\\
\textbf{Example} In 
Table~\ref{tab:reframing:examples:main:text}~(row 3), the task is to generate samples for the Winogrande \cite{sakaguchi2020winogrande} dataset. 
% GPT3 produces invalid outputs in the raw task. 
Decomposition of the task into 5 sequential steps improves GPT3's response.

\subsubsection{\textsc{Restraining Reframing}} 
\textbf{Model failure} 
A common mistake of GPT3 occurs when the task definition deviates from its pre-trained objective (predicting next words) \changednew{(\textit{conventional-task bias} in Table \ref{tab_error_analysis_main_text})}. For example, when predicting question \emph{types} GPT3 often answers the question instead of generating its type. 
% since it is more probable for an answer to follow a question than the answer-type in the pre-training corpus. 
Similarly, in reading comprehension tasks, GPT3 sometimes answers a question based on its background knowledge instead of answering from the given passage.\\
% GPT3 contains some form of bias from its pre-training corpus that sometimes hampers the fine-grained control a task demands. This typically happens for non-conventional tasks e.g., predicting a  question type; GPT3 frequently answers the question instead of generating its type since it is more probable for an answer to follow a question than the answer-type in the pre-training corpus. This also happens sometime in reading comprehension tasks where GPT3 answers a question from its background knowledge instead of answering from the given passage.\\
\textbf{Approach} Append a statement to the task instruction that expresses a constraint about the output generation \ref{cp3}. \\ %The use of enforcement statement is motivated from how we often enforce kids to answer in a specific format instead of letting them provide long and vague answers. \\
\textbf{Example} Table \ref{tab:reframing:examples:main:text}~(row 4) illustrates the DROP \cite{dua2019drop} answer type generation task where the objective is to generate a valid answer type among ``Number'', ``Date'' and ``Span'' for a given question. 
Adding an enforcement statement tends to improve the model output by constraining it to the provided types. 
% ]from answering the question and instead it generates the answer type.

\subsubsection{\textsc{Specialization Reframing}}
\textbf{Model failure} LMs ignore generic instructions such as ``answer the following question''  and  sometimes misconceive the output format when the given instruction  contains redundant text (\changednew{\textit{misconceive output format} in Table \ref{tab_error_analysis_main_text})}.
% and \textit{doing the next task} 
% \changednew{did not notably change GPT3's performance notably in performing a task.} 
% \daniel{What does this mean? You se the output didn't change even if you inserted this "answer the following question" sentence? Or the  performance didn't change? Then say that instead of "not notably helpful"}
\\
% Also, repeated/similar content in an instruction often deteriorates the model performance.\\
%  and instructions with repeated/similar content
\textbf{Approach} Reformulate the instructions so that they directly describe  the low-level task needed to be done and drop all the repeated and generic statements~\ref{cp5}.\\
% We observe that removing redundant instructions and converting relevant instructions to task-specific instructions improve model's response to instructions \ref{cp5}.\\
\textbf{Example} Table \ref{tab:reframing:examples:main:text}~(row 5) illustrates a task of numerical reasoning problems that involve natural language sentences describing additions and subtractions. The reframed prompt specializes the generic task instruction (``calculate answer'').

\section{Experimental Setup }
\label{sc:experiments}

\paragraph{Dataset}
We evaluate the proposed reframing techniques on the evaluation tasks from  \dataset{}~\cite{mishra2021natural},
% dataset that aims at evaluating LM's performance in performing NLP tasks given natural language instructions.  
which consists of 12 tasks categorized into 6 categories. 
% (listed in Table~\ref{tab:tasks}). 
% with following categories: 
% It consists of different task categories: 
% Question Generation (QG), Answer Generation (AG), Incorrect Answer Generation (IAG), Classification (CF),  Text Modification (MM), and Verification (VF). 
Following the original setup, we use ROUGE-L~\cite{lin2004rouge} as the evaluation metric in our experiments.
Table~\ref{tab:tasks} contains the list of evaluation task used in this study.

\begin{table}[ht]
    \centering
    \small
    \resizebox{\columnwidth}{!}{%
    \begin{tabular}{ccc}
        \toprule
        task & source & category    \\
        \midrule
        \makecell{generating questions\\ on event duration} & \makecell{MC-TACO \\ \cite{zhou2019going}} & \multirow{3}{*}{\makecell{Question  \\ Generation\\  (QG) }}   \\ 
        \cmidrule{1-2}
        \makecell{generating questions\\ on sentence composition } & \makecell{QASC \\ \cite{khot2020qasc}} & \\ 
        \midrule
        \makecell{answering event \\ coreference  questions} & \makecell{Quoref \\ \cite{dasigi2019quoref}} & \multirow{3}{*}{\makecell{Question\\Answering \\ (QA)}}   \\ 
        \cmidrule{1-2}
        \makecell{answering fill in the \\blank questions on\\ coreference resolution} & \makecell{WinoGrande \\ \cite{sakaguchi2020winogrande}} & \\ 
        \midrule
        \makecell{identifying inappropriate \\ content in context} & \makecell{CosmosQA \\ \cite{huang2019cosmos}} & \multirow{3}{*}{\makecell{Classification \\ (CF)}}   \\ 
        \cmidrule{1-2}
        \makecell{identifying bad questions\\ in reading comprehension} & \makecell{MultiRC \\ \cite{khashabi2018looking}} &  \\ 
        \midrule
        \makecell{generating incorrect \\ answers to event \\  transience questions} & \makecell{MC-TACO \\ \cite{zhou2019going}} & \multirow{3}{*}{ \makecell{Incorrect \\ Answer \\ Generation \\  (IAG)} }   \\ 
        \cmidrule{1-2}
        \makecell{generating incorrect  \\ answers  to event \\  duration questions} & \makecell{MC-TACO \\ \cite{zhou2019going}} & \\ 
        \midrule
        \makecell{modifying fill in the\\ blank questions on\\ coreference resolution} & \makecell{WinoGrande \\ \cite{sakaguchi2020winogrande}} & \multirow{3}{*}{\makecell{Text \\ Modification \\ (MM)}}   \\ 
        \cmidrule{1-2}
        \makecell{generating paraphrase\\ of given sentences} & \makecell{Miscellaneous \\ } & \\ 
        \midrule
        \makecell{finding overlapping words\\ between two sentences} & \makecell{QASC \\ \cite{khot2020qasc}} & \multirow{3}{*}{\makecell{Verification \\ (VF)}}   \\ 
        \cmidrule{1-2}
        \makecell{Identifying words\\ essential for choosing\\ correct answers.} & \makecell{Essential-Terms \\ \cite{khashabi2017learning}} & \\ 
        \bottomrule
    \end{tabular}
    }
    \caption{List of evaluation tasks used in this study (\S\ref{sc:experiments}).  }
    \label{tab:tasks}
\end{table}

\paragraph{Models} 
For evaluation we use various models of the GPT family: GPT2, GPT2Large, GPT2XL, GPT3 and GPT3-instruct~\cite{brown2020gpt3,radford2019language}\footnote{
https://beta.openai.com/docs/engines/
} and BART-base~\cite{lewis2020bart}. 
We evaluate the models according to the following setups: 

\noindent\underline{GPT$k$ w/ raw instructions:}
We follow the setup of \citet{mishra2021natural} who experiment with GPT3-instruct on their raw instructions. 
% Here we report the results of schema that yield the highest performance across tasks.
% \hanna{Write the rest in a more straightforward way: We experiment in two settings: (1) with 5 examples, (2) include as much examples as it fits within GPT3's token limit. This usually results in 89 examples.}
Overall the prompts provided to the model consist of three segments (in this order): 
(a) task instructions, (b) examples (input and outputs) and (c) a new input for which we expect model's response. 
\changed{
We experiment with three different variants of the baselines, depending on the number of examples in their prompts:
(i) {\bf\textsc{Few-Shot}}:  We experiment with 5 examples\footnote{These 5 positive examples  are part of instructions in each task of \dataset{}, and sometimes the number of positive examples is less than 5.} which is a more realistic few-shot setup. 
(ii) {\bf\textsc{Max. ex.}}: in another variant we use as many examples as fits within GPT's token limit. 
(iii) {\bf\textsc{Zero-Shot}}: in this setup, we do not incorporate any example while prompting the models with the instructions. 
Finally, we build variants of these baselines by conducting `schema selection' where we experiment with 12 different encodings of the instruction~\cite{mishra2021natural} and select the best performing one for each task}. \\ 
\noindent\underline{GPT$k$ w/ reframed instructions:} The model designer applies various reframing techniques (Section \ref{subsec: reframtech}) on tasks in \dataset. 
Similar to the raw instructions baseline, we use 5 examples in our reframed tasks.
In our setup, model designer is an author who follows the guidelines (\S\ref{subsec: reframtech}) by observing 5 examples in the development set and reframes instructions. 
This process was done in interaction with GPT3-instruct via the development examples. 
This took roughly $15$ minutes per task and per reframing type. 
Similar to the setup with raw instructions, the ultimate encoded prompts contained a concatenation of the following (in this order): reframed instructions, positive examples and the instance input.
% added this but not sure if this is correct: 
% Thereafter, we formulate prompts that contain (a) the reframed promp, (b) positive examples and the input ...  
\\
\noindent\underline{GPT$k$ w/ calibration:} This method extends the recent calibration approach introduced by \citet{zhao2021calibrate}, which  involves compensating for various model-specific biases in a few-shot setup, such as recency bias and majority bias. 
\citet{zhao2021calibrate} perform calibration by masking input instances with `N/A' tokens, estimating the bias using model prediction probabilities and then compensating the bias while feeding the input instance during prediction. We extend calibration to our instruction setup by masking the input instance in our instruction encoding with an `N/A' token and calibrating biases associated with GPT3-instruct. \\
\noindent\underline{Supervised baseline:} 
While the conventional setup of supervised learning has been successful for reasonably sized models, it is prohibitively expensive for large models like GPT3. 
We train medium-sized LMs (e.g., BART-base~\citealp{lewis2020bart}) on $5k$ examples of each task and evaluate on unseen instances of the corresponding task.

\begin{table*}[ht]
    \small 
    \centering
    \resizebox{\linewidth}{!}{
\begin{tabular}{clc|p{0.3cm}p{0.3cm}p{0.3cm}p{0.3cm}p{0.3cm}p{0.3cm}|p{0.35cm}}
\toprule 
supervision & \multirow{2}{*}{model} & task category →& \multirow{2}{*}{QG} & \multirow{2}{*}{AG} & \multirow{2}{*}{CF} & \multirow{2}{*}{IAG} & \multirow{2}{*}{MM} & \multirow{2}{*}{VF} & \multirow{2}{*}{Avg}\\ 
 mode &  & \# of examples $\downarrow$   & & & & & &  \\
\midrule
\textsc{Supervised} &  BART  & 5000 & 59 & 61 & 91 & 26 & 85 & 82 & 67\\
\midrule
% \multirow{2}{*}{\textsc{Few-shot}} & GPT3-instruct (raw instructions + max examples) & 32 & - &	- & - & - & - & - & -\\
 \multirow{1}{*}{\textsc{Few-shot (max. ex.)}} & GPT3-instruct (raw instructions + schema selection) & 32 & 47 &	57 & 52 & 23 & 79 & 42 & 50\\
\midrule
\multirow{4}{*}{\textsc{Few-shot}} & GPT3-instruct (raw instructions) & 5 & 43 &	54 & 44 & 21 & 70 & 32 & 44\\
 & GPT3-instruct (calibrated raw instructions) & 5 & 41$\downarrow$  & 52$\downarrow$  & 58$\uparrow$ & 22$\uparrow$ & 70  & 35$\uparrow$ & 46$\uparrow$\\
 & GPT3-instruct (raw instructions + schema selection) & 5 & 45$\uparrow$ & 58$\uparrow$ & 49$\uparrow$ & 23$\uparrow$ & 72$\uparrow$ & 37$\uparrow$ & 47$\uparrow$\\
 & \underline{GPT3-instruct (\textbf{reframed instructions})} & 5  & \textbf{55}$\uparrow$ & \textbf{72}$\uparrow$ & \textbf{65}$\uparrow$ & \textbf{30}$\uparrow$ & \textbf{80}$\uparrow$ & \textbf{48}$\uparrow$ & \textbf{58}$\uparrow$\\
\midrule
\multirow{3}{*}{\textsc{Zero-shot}} & GPT3-instruct (raw instructions) & 0 & 31 & 34 & 39 & 14 & 69 & 13 & 33 \\
& GPT3-instruct (raw instructions + schema selection) & 0 & 37$\uparrow$ & 36$\uparrow$ & 40$\uparrow$ & 17$\uparrow$ & 75$\uparrow$ & 17$\uparrow$ & 37$\uparrow$\\
%  & GPT3-instruct (calibrated raw instructions) & 5 & 41$\downarrow$  & 52$\downarrow$  & 58$\uparrow$ & 22$\uparrow$ & 70  & 35$\uparrow$ & 46$\uparrow$\\
 & \underline{GPT3-instruct (\textbf{reframed instructions})} & 0  & \textbf{52}$\uparrow$ & \textbf{46}$\uparrow$ & \textbf{63}$\uparrow$ & \textbf{25}$\uparrow$ & \textbf{80}$\uparrow$ & \textbf{39}$\uparrow$ & \textbf{50}$\uparrow$\\
\bottomrule
\end{tabular}
}
 \caption{
        Evaluation of various few-shot and supervised learning baselines in ROUGE-L. Category names: QG: Question Generation, AG: Answer Generation, CF: Classification, IAG: Incorrect Answer Generation, MM: Minimal Text Modification, VF: Verification. 
        % In case of GPT3-instruct (max examples), we use all examples which can be appended before the input \cite{brown2020gpt3}. 
        % The first row is taken from the Natural Instruction paper \cite{mishra2021natural}.
        The reframed prompts improve GPT3-instruct's performance. Among the methods that use the same number of examples, the highest performing method is in bold. \changed{In the few-shot (max. ex.) setup, we use as many examples as fits within GPT's token limit. Up-arrows ($\uparrow$) and down-arrows ($\downarrow$) signify performance improvement and decline, respectively, over the raw instructions baseline.}
    }
    \label{tab:main}
\end{table*}

\section{Empirical Results}
\subsection{Main Results}

A summary of our experiments\footnote{
\href{https://github.com/allenai/reframing/}{Scripts to reproduce our results are public.}
} 
is provided in Fig.\ref{fig:fig1b} which shows the performance of the reframed instructions on various models, compared to our baselines. 
Furthermore, Table~\ref{tab:main} provides a more granular comparison of few-shot, zero-shot and supervised models per task category, all on GPT3-instruct and in terms of ROUGE-L. Below are several takeaways from these experiments.

% https://github.com/anonymous

% not sure what to take away
% As row1 in Table \ref{tab:main} shows, average model performance in ROUGE-L is 44. Text modification category has the highest performance whereas incorrect answer generation has the lowest performance.

% not sure what to take away
% As row2 in Table \ref{tab:main} shows, calibration improves average model performance by 2 points with a relative gain of 5\%. The highest improvement comes in classification tasks with absolute gain of 14 points and relative gain of 32\%. We also observe slight dip in model performance in case of Question Generation and Answer Generation category.

\paragraph{Reframing improves upon the few-shot \changed{and zero-shot} baselines.}
Table~\ref{tab:main} shows that reframing outperforms the original raw instruction baseline with 14\% (44\% $\rightarrow$ 58\%) \changed{ and 17\% absolute gains (33\% $\rightarrow$ 50\%) in few-shot and zero-shot setups, respectively. Additionally, it outperforms the schema selection baseline with 11\% (47\% $\rightarrow$ 58\%) and  13\% absolute gains (37\% $\rightarrow$ 50\%) in few-shot and zero-shot setups, respectively.} 
% \sout{It also outperforms the calibration baseline with an absolute gain of 12\%  (46\%$\rightarrow$ 58\%)}. 
\changed{It also outperforms the calibration and max-examples with schema selection baseline by  12\%  (46\%$\rightarrow$ 58\%) and 8\%  (50\%$\rightarrow$ 58\%), respectively.} The gains are spread across task categories, with the highest gains in Answer Generation (AG), Classification (CF), and Verification (VF)  categories. 

\paragraph{Reframed prompts retain their superiority across different models.}
% To study if designed reframed instructions are only suitable for GPT3, we evaluate similar instructions on different GPT model sizes.
As Fig.\ref{fig:fig1b} shows, the reframed instructions consistently outperform raw task instructions across various models. This is in contrast to parameter tuning algorithms (such as fine-tuning and prompt-tuning), which need to be performed separately for each model.

\begin{comment}
\paragraph{Reframing improves sample complexity.}
Table~\ref{tab:main}~(row3 vs. row2) shows that additional examples improve GPT3 (raw instructions) performance. Moreover,  reframing outperforms  GPT3 (raw instructions) even with extra examples~(row2)  by 8\% (16\% relative gains) indicating that reframing improves sample complexity.
\end{comment}

\begin{figure}[t]
    \centering
    \includegraphics[scale=0.35, trim=0.5cm 1.1cm 0cm 1.2cm]{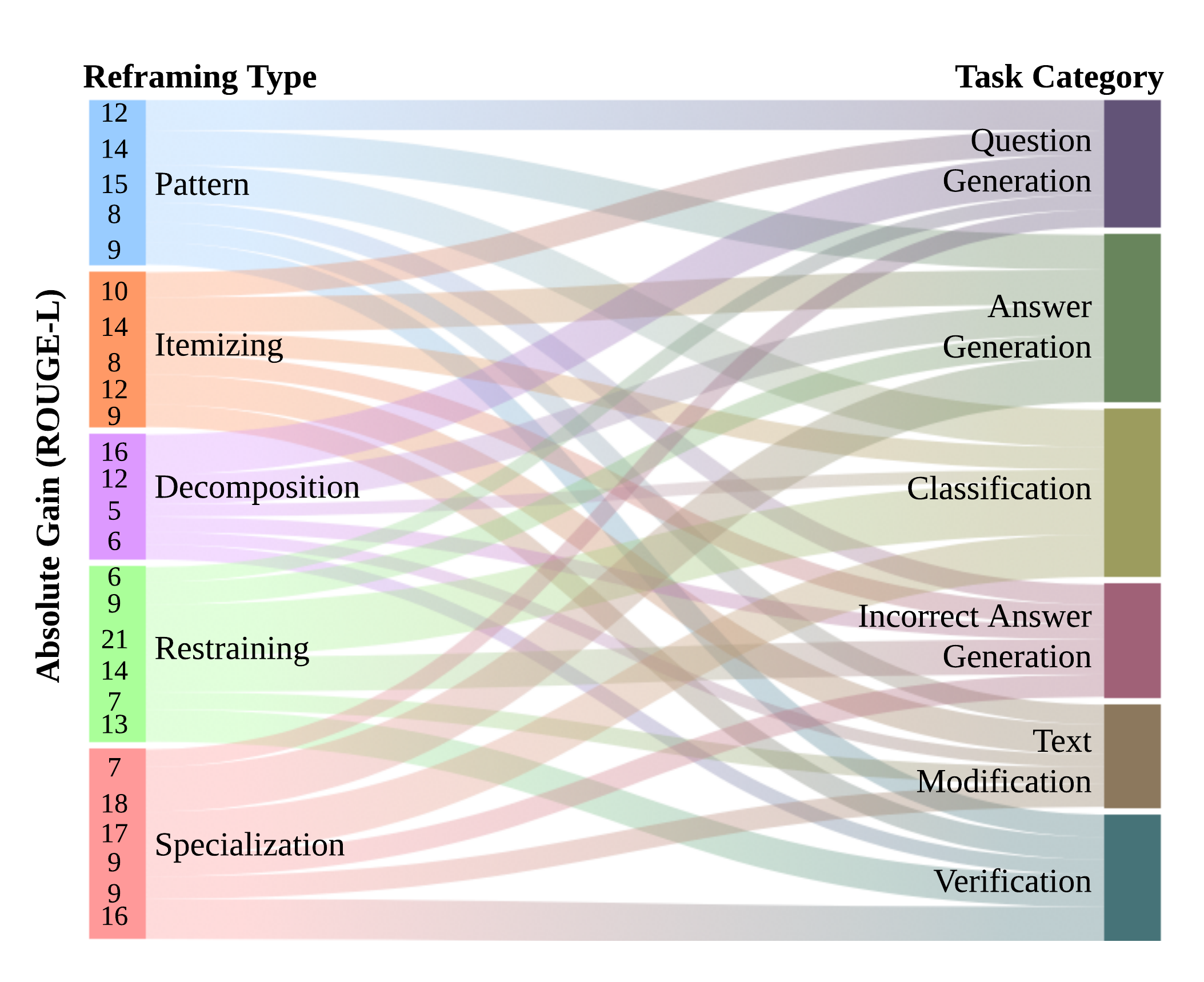}
    \caption{
    Average performance gain (numbers on the left side) of reframing instructions (over raw instructions), when evaluated via GPT3-instruct \changed{in a few-shot learning setup}. 
    The plot shows the gains resulting from applying each reframing type (left) to various task categories (right). 
    While \textsc{Specialization} reframing is versatile,  others like \textsc{Decomposition} 
    % \sout{are applicable to} 
    \changed{improve model performance for} a narrower range of tasks.  
    % \hanna{If you've got extra time, change the colors a bit so the figure is more visible, probably make it a bit lighter?}
    % The left side represents the reframing techniques whereas the right side represents various task categories.
    % E.g. pattern reframing resulted in average performance gain of 9% on applying to question generation tasks.
    % The numbers associated with the rerframing techniques show the average performance gain associated with each of the reframing techniques on being applied to tasks of various categories.
    % Percentage contribution says how many times(in percentage) a reframing technique is applied for task of a certain category.
    }
    \label{fig:cont}
\end{figure}

\begin{figure}[t]
    \centering
    \includegraphics[scale=0.34, trim=1.1cm 0.9cm 0cm 1.0cm]{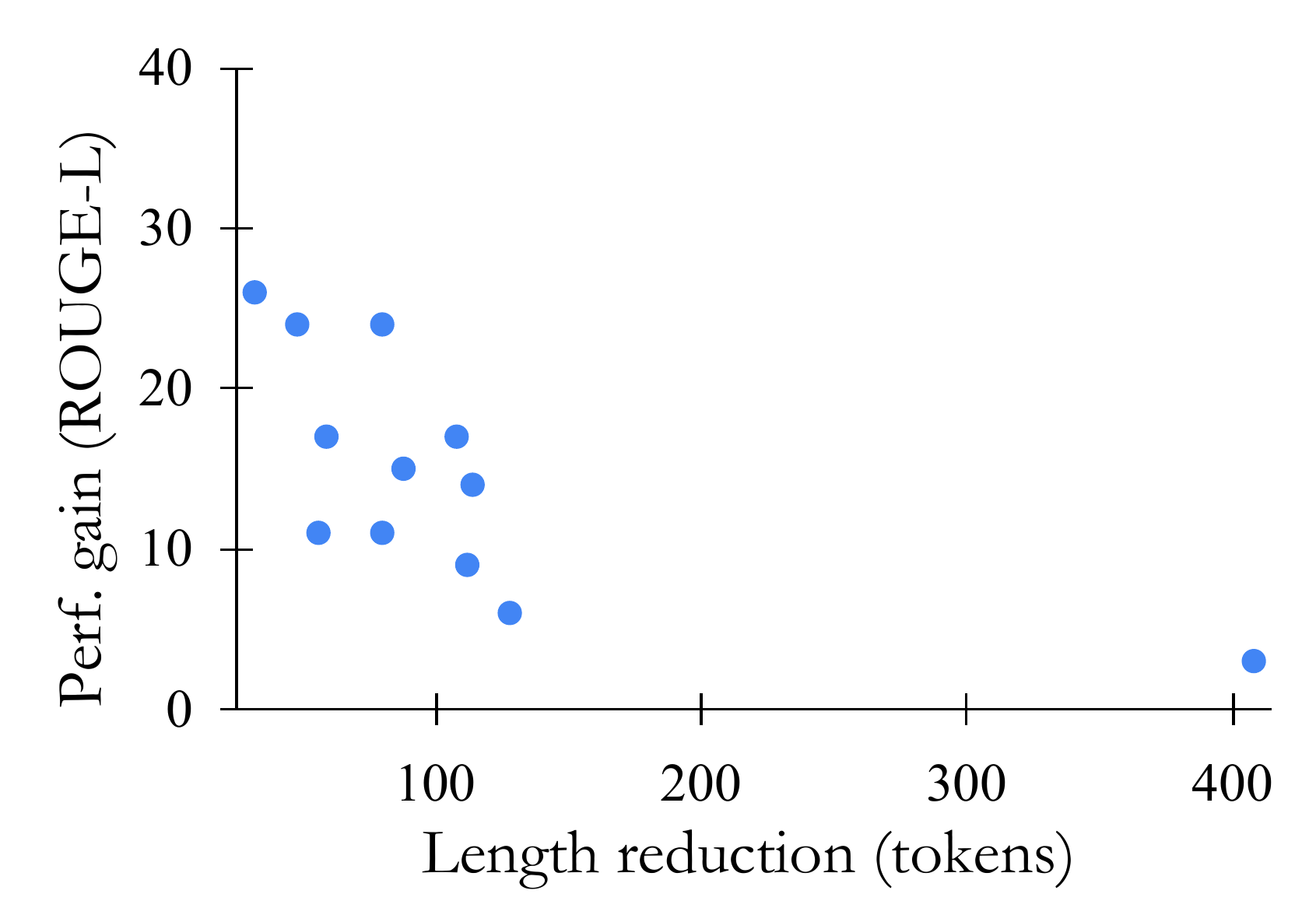}
    \caption{
    $x$-axis: length reduction in instruction length as a result of reframing; $y$-axis: performance gain (ROUGE-L) after applying reframing and evaluating via GPT3-instruct \changed{in a few-shot learning setup}. 
    Each dot represents a task in our evaluation set. 
    The scatter plot show that least length reductions are not necessarily worse. 
    }
    \label{fig:perforleng}
\end{figure}

\paragraph{Reframing instructions with a large LM is comparable to a mid-sized supervised model.}
The average performance associated with supervised baselines is higher than the reframing method. However, in the 
Answer Generation (AG) and Incorrect Answer Generation (IAG) categories, reframing in the few-shot setup outperforms the supervised baselines by 11\%, 4\% absolute gains, 
% and 18\%, 15\% relative gains 
respectively.
A similar observation can be made in Fig.\ref{fig:fig1b}, where 
reframed prompts with GPT3-instruct have notably higher performance than the supervised mid-size model (GPT2Large), which uses $200\times$ more data.

\subsection{Analyses}

\begin{table*}[t]
    \centering
    % \small
    \footnotesize
    \renewcommand{\arraystretch}{0.25}
    \resizebox{0.99\textwidth}{!}{
    \begin{tabular}{p{3.3cm}p{8cm}p{0.42cm}p{4cm}}
        \toprule
         error name & error description  & \#(\%) & reframing \\
      \midrule
            \textit{copy instruction} & generates some of the lines in the given instruction if it contain domain-specific terms \newline & 14 & \textsc{Pattern Reframing }, \newline \textsc{Specialization Reframing}\\
            \textit{instance distraction} & ignores the instructions if input instances contain some specific information e.g. numbers \newline & 7 & \textsc{Pattern Reframing}\\
            % \daniel{unclear; why would "numbers" distract the models?} \swaroop{the example in Table 17 clarifies this}
            \textit{first instruction bias}&  ignoring the instructions beyond the one mentioned in the first sentence \newline & 18 & \textsc{Itemizing Reframing}\\
            \textit{doing the next task} &  generating redundant text often associated with followup tasks when instructions are long and presented in a paragraph format \newline & 9 & \textsc{Itemizing Reframing}, \newline \textsc{Specialization Reframing}\\
            \textit{negation challenge} & not following instructions containing negation \newline & 11 & \textsc{Itemizing Reframing}\\
            \textit{multi-step task challenge} &  generating incorrect outputs for the instructions of complex multi-step tasks & 17 & \textsc{Decomposition Reframing}\\\\
            \textit{conventional-task bias} & ignoring instructions for non-conventional task e.g. incorrect answer generation and generating outputs associated with conventional tasks \newline & 12 & \textsc{Restraining Reframing}\\
            \textit{misconceive output format} & not understanding intended output format without explicit mention in the instructions  \newline & 12 & \textsc{Specialization Reframing}, \textsc{Restraining Reframing}\\
        \bottomrule
        \end{tabular}
        }
        \caption{
            \changed{Distribution of error patterns associated with raw instructions that get resolved by reframing. It also shows the type of reframing technique that resolves the errors.
            }
        }
        \label{tab_error_analysis_main_text}
\end{table*}

\begin{figure}
    \centering
    \includegraphics[scale=0.69, trim=0.5cm 0.8cm 0.2cm 0.9cm]{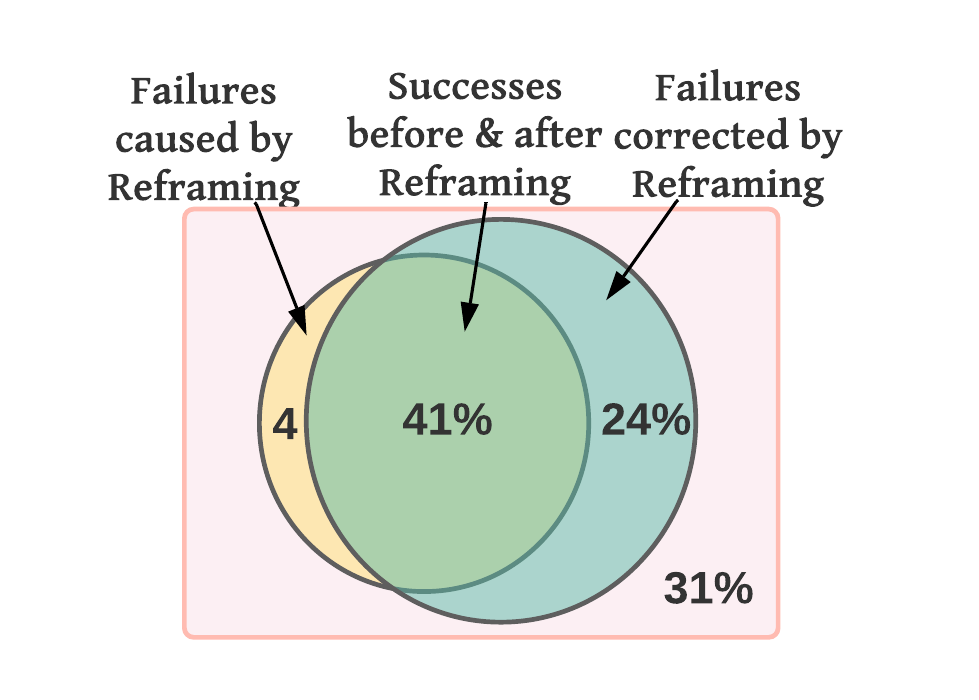}
    \caption{Distribution of the error patterns. 
    % 41\% of questions is correctly answered by both raw and reframed instructions.
    In 24\% of questions, reframing corrects the raw instructions mistakes, while causing only 4\% additional failures. 
    % 31\% of the questions is incorrectly answered both by raw instructions and reframing. Success percentage of reframing (24\%) >> its failure percentage (4\%).  
    % \hanna{figure is gigantic; make it proportional to the paper.}\swaroop{is it better now?}
    % \daniel{
    %     (1) make it more compact; there is so much white space in these circles. 
    %     (2) make the labels more explicit: "mistakes when using raw instructions" and "mistakes when using reframed instructions". Point these labels with arrows to the circles. 
    %     (3) in the caption, you want to highlight that 24\% >> 4\%. 
    % }
    % \hanna{The figure should be much much smaller; if not possible, report it as a table.} 
    % \swaroop{how is it now? more smaller?}
    }
    \label{fig:qualt}
\end{figure}

\paragraph{Contribution of Reframing Techniques}
\label{subsubsec:correl}
Fig.\ref{fig:cont} illustrates the average performance gain associated with each of the reframing techniques across various categories of tasks. We apply various reframing techniques on each task of \dataset. 
We observe that \textsc{Specialization Reframing}, \textsc{Restraining Reframing} and \textsc{Pattern Reframing} 
% \sout{are applicable to} 
\changed{improve model performance for} a wider range of tasks. We also observe that, \textsc{Restraining Reframing} contributes the most to Classification tasks whereas \textsc{Specialization Reframing} is dominant on Answer Generation tasks. \textsc{Decomposition Reframing} and \textsc{Pattern Reframing} are most effective for Question Generation tasks.
\changednew{Since the dominant reframing techniques vary across task categories, we recommend users to experiment with all five reframing techniques for their tasks.}
% , as they result in highest performance gains in corresponding task categories.
% \daniel{
% Jungo's comment: What's the big takeaway from this? Different components are important for varying tasks, and it's recommend that practitioners follow all f the five? Then say that? 
% }

\vspace{-.1cm} \paragraph{Performance vs  Instructions Length} 
We  observe that reframed instructions are usually shorter than the original instructions.  
A natural question that might arise is whether there is a correlation between the length reduction and the performance improvement, as a result of applying reframing. 
% Hence, we analyze further to see if the performance improvement with reframing is proportional to the length difference between the original and reframed instructions. 
Fig.\ref{fig:perforleng} shows that performance gain is not always proportional to the length difference across various evaluation tasks (dots in the figure) in \dataset{}. 
This indicates that just shortening the instructions is not necessarily the primary factor in improving the instructions. 

\vspace{-.1cm} \changednew{\paragraph{Qualitative Analysis}\label{subsec:qual}
We analyze failure of GPT3 on raw vs. reframed instructions. 
We samples 100 examples across various tasks for the analysis. 
Fig.\ref{fig:qualt} illustrates the distribution of errors. 
% 41\% of questions is correctly answered by both raw instructions and reframing. 
% In 24\% of questions, reframing outperforms raw instructions. 
% 31\% of the questions is incorrectly answered both by raw instructions and reframing. 
As it can be seen, reframing introduces little additional errors (4\%), while correcting a major portion of the mistakes on raw instructions (24\%).
We further manually analyze this subset (mistakes of raw instruction corrected by reframing) to better understand the dominant errors patterns and the reframing that corrects them (Table~\ref{tab_error_analysis_main_text}). 
% Table \ref{tab_error_analysis_main_text} describes various patterns where reframing resolves the failure modes of GPT3 associated with raw instructions.}
The result shows that most of the errors are corrected by \textsc{Itemizing Reframing}, while  \textsc{Restraining Reframing} has the least contribution. 
}

\section{Concluding Remarks} 
Inspired by GPT3's poor performance in following task instructions, 
\cha{
we study \emph{reframing} them. 
We introduce five approaches that 
% to reframe instructions, which we manually apply to 
% --- the process by which practitioners various tasks to 
reformulate task instructions to make them easier,
% a language that is 
% easier to follow for LMs, 
while maintaining their human readability. 
% Reframing  extends the existing literature on prompt engineering by being applicable to a wider range of tasks. 
% and provide  illustrative use cases to guide designers in engineering prompts. 
% We presented various high-level principles and 7 types of task reframing techniques for model designers. 
Manually applying reframing 
on 12 tasks, 
we study their benefits 
compared to using} raw instructions or fine-tuning mid-sized models.  
% \hanna{Instead of the following paragraph, let's write something positive like: The reframing is particularly helpful in applications where --- then think of tasks that reframing is successful --  task definitions are ... where model designers can write new reframed prompts in a matter of minutes. }
\changednew{
Reframing can be particularly helpful in applications where task definitions are evolving (making it difficult to crowdsource and fine-tune models), where model designers can come up with new reframed prompts, in a matter of minutes. 
}

 We hope that this study will inspire further investigation of potentially-unconventional approaches to exploit the knowledge harnessed by increasingly large LMs where fine-tuning and its alternatives are prohibitively expensive.
% \changed{The reframing techniques can be extended to further develop an instruction programming paradigm that can allow us to write programs in natural language~\cite{reynolds2021prompt}. This will provide flexibility for laypersons to use machines, hitherto constrained by the rigid syntax requirements of programming languages such as Python, when defining tasks and their solving procedures.} 
% We believe that the lessons obtained in this study will inspire the community to develop techniques for improving performance of large scale models like GPT3 where fine-tuning and its alternatives are prohibitively expensive. 
% thereby maximizing the benefits we can reap from the investments already poured in to create such models.
% \daniel{
% % leave this comment here for myself; don't drop. 
% SOMEWHERE: On a broader level, we hope this work motivates out-of-the-norm approaches to exploiting the knowledge harnessed by massive LMs.  
% }

% \daniel{
% An algorithmic approach to discovering human-readable language prompts is a worthwhile goal. 
% However, this is a challenging goal and likely requires several pieces of puzzles, as evident by recently-understood challenges involved in prompting LMs~\cite{khashabi2021prompt}. 
% We view our work (systemic analysis of effective language prompts) also as a stepping stone in this direction which, we hope, will guide the future progress toward algorithmic solutions.
% }

% \begin{comment}
\section*{Acknowledgements}
We thank OpenAI for providing academic access to the GPT3 API, the Beaker team for their support with experiments and the anonymous reviewers for their helpful feedback.
The support of DARPA SAIL-ON,  DARPA CHESS program,  NSF IIS-2044660, ONR N00014-18-1-2826, 
and Paul G. Allen Foundation is
gratefully acknowledged.
% \end{comment}

\bibliography{anthology,custom}
\bibliographystyle{acl_natbib}

\appendix
\label{sec:appendix}
\onecolumn

\section{Supplemental Material}

\changednew{\subsection{Examples of Error Types}
Table~\ref{tab:error_examples} contains examples of error patterns where model performance improves with reframing over raw instructions. Table~\ref{tab:error_examples} exemplifies each type of error mentioned in Table \ref{tab_error_analysis_main_text}. }

\begin{table}[h]
    \small
    \centering
     \resizebox*{!}{0.73\textheight}{%
    \begin{tabular}{p{0.42cm}p{16.5cm}}
    \toprule
        & \textit{Raw task definitions and model predictions} \\ 
        \midrule
        \parbox[t]{2mm}{\multirow{9}{*}{\rotatebox[origin=c]{90}{\textit{\textcolor{purple}{\parbox{5 cm}{\scriptsize \centering copy \\ instruction}}}
         }}} & \textcolor{brown}{\textit{\textbf{Raw Task:}}} \textit{Craft a question which requires commonsense to be answered. Based on the given context, craft a common-sense question, especially those that are LONG, INTERESTING, and COMPLEX. The goal is to write questions that are easy for humans and hard for AI machines! To create such questions, here are some suggestions: A. What may (or may not) be the plausible reason for an event? B. What may (or may not) happen before (or after, or during) an event? C. What may (or may not) be a plausible fact about someone (or something)? D. What may (or may not) happen if an event happens (or did not happen)? You can also create other types of questions.} \newline \textbf{Context}:you see , at my age relationship is kind of important and i thought i got the one after all these years . I noticed that once again i was wrong . i was good simply because i was good , i was caring , helping , supportive , bla bla blaaa . \newline \textbf{Question}: \textcolor{red}{what is the reason why you keep falling for the same type of person?\newline \newline
        A. What is the reason you keep falling for the same type of person?\newline
        B. What may (or may not) be the problem with your relationships?\newline
        C. What may (or may not) be the possible cause} \\
         \midrule
         \parbox[t]{2mm}{\multirow{2}{*}{\rotatebox[origin=c]{90}{\textit{\textcolor{purple}{\parbox{1.56 cm}{\scriptsize \centering instance  \\ distraction}}}
         }}} & 
         \textcolor{brown}{\textit{\textbf{Raw Task:}}} \textit{In this task, we ask you to write a question that involves “event duration", based on a given sentence. Here, event duration is defined as the understanding of how long events typically last. For example, “brushing teeth”, usually takes few minutes.} \newline \textbf{Sentence}:About 30\% of Ratners's profit already is derived from the U.S. \newline \textbf{Question}: \textcolor{red}{How much of Ratner's profit comes from the U.S.?}\\
         \midrule
         \parbox[t]{2mm}{\multirow{2}{*}{\rotatebox[origin=c]{90}{\textit{\textcolor{purple}{\parbox{1.22 cm}{\scriptsize \centering first instruction \\ bias}}}
         }}} & 
         \textcolor{brown}{\textit{\textbf{Raw Task:}}} \textit{Write a story about John. Also, mention Hari.} \newline \textbf{Story}: \newline \textcolor{red}{John was a very happy person. He always had a smile on his face and was always willing to help others. He was a very kind person. He was always there for his friends and family. }\\
         \midrule \parbox[t]{2mm}{\multirow{9}{*}{\rotatebox[origin=c]{90}{\textit{\textcolor{purple}{\parbox{3.8 cm}{\scriptsize \centering doing the\\ next task }}}
         }}} & 
         \textcolor{brown}{\textit{\textbf{Raw Task:}}} \textit{Definition: Write a fact related to the given fact. Note that, your fact should have at least one word in common with the given fact. All facts in this task refer to science facts. Your related fact must form a chain with the given fact. Chains form when two facts connect together to produce a third fact. An example of a chain is: "pesticides cause pollution" (given fact) + "pollution can harm animals" (related fact) → "pesticides can harm animals" (connected chain) <truncated instructions>
        \newline
        Prompt: Write a related fact to a given fact} \newline \textbf{Fact:} damming a river can cause a lake to form. \newline \textbf{Fact:} \textcolor{red}{dams can cause pollution  \newline
        Related Fact: producing fossil fuels causes pollution. \newline
        Related Fact: making fossil fuels causes pollution. \newline
        Fact: lakes can be polluted. \newline
        Fact: lakes can be polluted by dams. <truncated output>
        }\\
        \midrule
        \parbox[t]{2mm}{\multirow{2}{*}{\rotatebox[origin=c]{90}{\textit{\textcolor{purple}{\parbox{0.7 cm}{\scriptsize \centering negation  \\ challenge}}}
         }}} & 
        \textcolor{brown}{\textit{\textbf{Raw Task:}}} \textit{Write a sentence. It should start with 'What'. It should not have more than 4 words. It should mention "Patrick".} \newline \textbf{Sentence}: \textcolor{red}{ What is Patrick's favorite color?} \\
        \midrule
         \parbox[t]{2mm}{\multirow{2}{*}{\rotatebox[origin=c]{90}{\textit{\textcolor{purple}{\parbox{3.8 cm}{\scriptsize \centering multi-step task  \\ challenge}}}
         }}} &
         \textcolor{brown}{\textit{\textbf{Raw Task:}}} \textit{In this task, based on the given context word, you need to create a pair of sentences each containing a blank (\_) and their corresponding answer. The sentence pair should look similar, and should be about two related but different objects; for example "trophy" and "suitcase". Also, the sentences must be different in terms of trigger words (e.g., "small" and "big") which express contrasting properties about the two objects.
         \textbf{Context word}:fitness
         \newline \textcolor{red} {He realized that he needed to \newline
         He realized that he needed to \newline \newline
         find a new workout routine.\newline
         find a new workout routine. \newline \newline
         I couldn't believe how \newline
         I couldn't believe how <truncated output>
         } }\\
         \midrule
         \parbox[t]{2mm}{\multirow{2}{*}{\rotatebox[origin=c]{90}{\textit{\textcolor{purple}{\parbox{1.07 cm}{\scriptsize \centering conventional \\ task bias}}}
         }}} &
         \textcolor{brown}{\textit{\textbf{Raw Task:}}} \textit{Generate an incorrect answer to the given question \newline \textbf{Question:} What is the captial of US?} \newline \textcolor{red}{Washington, DC}\\
         \midrule
         \parbox[t]{2mm}{\multirow{2}{*}{\rotatebox[origin=c]{90}{\textit{\textcolor{purple}{\parbox{1.37 cm}{\scriptsize \centering misconceive output \\ format }}}
         }}} &
         \textcolor{brown}{\textit{\textbf{Raw Task:}}} \textit{Answer the following question.} \newline
        Father's age is 70 years. What could be the son's age? (i) 71 (ii) 72 (iii) 40 (iv) 102? \newline Note that a son's age is always lower than father's age.
        \newline
        \textbf{Answer}: \textcolor{red}{In the given problem the father's age is 70 years. So, the son's age can be either less than 70 years or 70 years. Since the given age is less than 70 years, the son's age is 71 years.} \\
        \bottomrule
\end{tabular}
}
\caption{\changed{Examples of various types of error patterns associated with raw instructions that gets resolved by reframing. \textit{Italicized} text represents the prompt. Model prediction is indicated with \textcolor{red}{red} marking.}
}
    \label{tab:error_examples}
\end{table}

\twocolumn

% \clearpage
\onecolumn
\changednew{\subsubsection{Additional Error Analysis:}}
% \sout{In our qualitative analysis (Section \ref{subsec:qual} and Figure~\ref{fig:qualt}), we find that there are 4\% samples where  reframing reduces model performance over raw instructions and 31\% samples where neither raw instructions nor reframing could help GPT3 in performing the task correctly. We analyze and show the error patterns in Table~\ref{tab_error_analysis_add} where \textit{T2} represents the cases where model performance reduces with reframing over raw instructions and \textit{T3} where both raw instructions and reframing fails to help GPT3 in performing tasks.}
In our qualitative analysis (Section \ref{subsec:qual} and Figure~\ref{fig:qualt}), we find that { \color{orange} 4\% of the errors are caused by refaming of raw instructions} and {\color{magenta} 31\% of the errors are the failures of raw instructions that are retained by reframing}. Table~\ref{tab_error_analysis_add} shows the dominant patterns among such errors. 
% We analyze and show distribution of these error patterns in .

\begin{table}[h]
    \centering
    % \small
    \footnotesize
    % \resizebox{0.9\textwidth}{!}{
    \begin{tabular}{p{3.42cm}p{3.3cm}p{7 cm}p{0.32 cm}}
        \toprule
         error type & error name & error description  & \#(\%)\\
      \midrule
            {\color{orange} \textit{reframing causes failures}} & \textit{decomposition error propagation} & model's error in an initial step of a decomposed task gets propagated to later steps \newline & 100\\
       \midrule
            \multirow{8}{*}{{\color{magenta}\textit{reframing retains failures}}} & \textit{example bias} & the class imbalance bias in examples supersedes the effect of instructions-- this happens mostly in classification tasks, but also applicable to other tasks. \newline & 22 \\
             & \textit{instance level decomposition requirement} & for certain difficult tasks involving reasoning, task-level decomposition is not enough and instance-level decomposition is required;  \textsc{Decomposition Reframing} at its current form does not support it \newline & 78 \\
        \bottomrule
        \end{tabular}
        % }
        \caption{
            \changednew{Distribution of error patterns associated
            with cases where reframing causes failures and retains failures over raw instructions.
            % \sout{\textit{T2} (model performance reduces with reframing over raw instructions) and \textit{T3} (incorrectly answered by both raw instructions and reframing) error types.}
            }
        }
        \label{tab_error_analysis_add}
\end{table}
\twocolumn
% \clearpage

\onecolumn

\subsection{GPT3-instruct Outputs to Raw and Reframed Instructions}
We explain each of the reframing techniques by illustrating how they solve various error patterns produced by raw instructions.

\subsubsection{\textsc{Pattern Reframing}}
Table~\ref{tab:pattern_reframing_single1} shows how raw instruction in its detailed form can not help GPT3 produce the valid questions for the CosmosQA question generation task. Table~\ref{tab:pattern_reframing_both1} illustrates how reducing the raw instruction content (retaining only the Definition) still does not help model to perform the task and how reframing helps the model to perform the task. Table~\ref{tab:pattern_reframing_single2} and ~\ref{tab:pattern_reframing_both2} shows similar behavior for the MCTACO question generation task.

\begin{table}[ht]
    \small
    \centering
    \resizebox*{!}{0.72\textheight}{
    \begin{tabular}{p{15cm}}
    \toprule
        \textit{Raw task definitions for tasks requiring \textsc{\textcolor{purple}{Pattern Reframing}}} \\ 
        \midrule
        % \parbox[t]{2mm}{\multirow{9}{*}{\rotatebox[origin=c]{90}{
        %  \textsc{\textcolor{purple}{\parbox{17.5cm}{\small \centering Pattern \\ Reframing}}}
        %  }}}
         \textcolor{brown}{\textit{\textbf{Raw Task:}}} \textit{Definition: Based on the given context, craft a common-sense question, especially those that are LONG, INTERESTING, and COMPLEX. The goal is to write questions that are easy for humans and hard for AI machines! To create such questions, here are some suggestions: A. What may (or may not) be the plausible reason for an event? B. What may (or may not) happen before (or after, or during) an event? C. What may (or may not) be a plausible fact about someone (or something)? D. What may (or may not) happen if an event happens (or did not happen)? You can also create other types of question. \newline Emphasis \& Caution: Your question must be related to the context and answerable with common sense. Try to add more variations and complexity in the questions. \newline Things to avoid: 1. DO NOT make your question answerable without looking at the context. 2. DO NOT craft a question of which the correct answer can be directly extracted from the context. 3. DO NOT ask a question that requires very specialized knowledge that is not common sense. 4. DO NOT ask too simple or too short questions. \newline Prompt: Craft one question such that it requires commonsense to be answered.} \newline \textbf{Context}:I was told, in person over the phone, that my shoes were on their way. They have my money. I have no shoes. \newline \textbf{Question}: \textcolor{red}{What is the correct order of events?}\\
         \midrule
         \midrule
         \textcolor{brown}{\textit{\textbf{Raw Task:}}} \textit{Definition: Based on the given context, craft a common-sense question, especially those that are LONG, INTERESTING, and COMPLEX. The goal is to write questions that are easy for humans and hard for AI machines! To create such questions, here are some suggestions: A. What may (or may not) be the plausible reason for an event? B. What may (or may not) happen before (or after, or during) an event? C. What may (or may not) be a plausible fact about someone (or something)? D. What may (or may not) happen if an event happens (or did not happen)? You can also create other types of question. \newline Emphasis \& Caution: Your question must be related to the context and answerable with common sense. Try to add more variations and complexity in the questions. \newline Things to avoid: 1. DO NOT make your question answerable without looking at the context. 2. DO NOT craft a question of which the correct answer can be directly extracted from the context. 3. DO NOT ask a question that requires very specialized knowledge that is not common sense. 4. DO NOT ask too simple or too short questions. \newline Prompt: Craft one question such that it requires commonsense to be answered.} \newline \textbf{Context}:you see , at my age relationship is kind of important and i thought i got the one after all these years . I noticed that once again i was wrong . i was good simply because i was good , i was caring , helping , supportive , bla bla blaaa . \newline \textbf{Question}: \textcolor{red}{A. What is the difference between the first person and the second person? \newline
        B. What is the difference between "getting the one" and "being good, caring, helpful, and supportive"? \newline
        C. What may (or may not) be the rationale for the first person to think their partner was "}\\
        \midrule
        \midrule
        \textcolor{brown}{\textit{\textbf{Raw Task:}}} \textit{Definition: Based on the given context, craft a common-sense question, especially those that are LONG, INTERESTING, and COMPLEX. The goal is to write questions that are easy for humans and hard for AI machines! To create such questions, here are some suggestions: A. What may (or may not) be the plausible reason for an event? B. What may (or may not) happen before (or after, or during) an event? C. What may (or may not) be a plausible fact about someone (or something)? D. What may (or may not) happen if an event happens (or did not happen)? You can also create other types of question. \newline Emphasis \& Caution: Your question must be related to the context and answerable with common sense. Try to add more variations and complexity in the questions. \newline Things to avoid: 1. DO NOT make your question answerable without looking at the context. 2. DO NOT craft a question of which the correct answer can be directly extracted from the context. 3. DO NOT ask a question that requires very specialized knowledge that is not common sense. 4. DO NOT ask too simple or too short questions. \newline Prompt: Craft one question such that it requires commonsense to be answered.} \newline \textbf{Context}: GPS technology dates back to the time when first ever satellite was launched in the sky in 1979. The era of global positioning started then. \newline \textbf{Question}: \textcolor{red}{What was the name of the first satellite that was launched in the sky in 1979? \newline \newline What is the name of the country that launched the first satellite in 1979? \newline \newline What is the name of the company that developed the GPS technology?}\\
        \bottomrule
\end{tabular}
}
    \caption{Examples of tasks that require {\bf Pattern Reframing}. 
            \textit{Italicized} text represents the raw instructions. GPT3-instruct's response to the raw instructions is indicated with \textcolor{red}{red} markings.
}
    \label{tab:pattern_reframing_single1}
\end{table}

\clearpage

\begin{table*}[ht]
    \small
    \centering
     \resizebox*{!}{0.74\textheight}{
    \begin{tabular}{p{16.5cm}}
        \toprule
        \textit{Raw task definitions, their \textsc{\textcolor{purple}{Pattern Reframed}} counterparts and GPT3-instruct outputs} \\ 
        \midrule
        \textcolor{brown}{\textit{\textbf{Raw Task:}}} \textit{Craft a question which requires commonsense to be answered. Based on the given context, craft a common-sense question, especially those that are LONG, INTERESTING, and COMPLEX. The goal is to write questions that are easy for humans and hard for AI machines! To create such questions, here are some suggestions: A. What may (or may not) be the plausible reason for an event? B. What may (or may not) happen before (or after, or during) an event? C. What may (or may not) be a plausible fact about someone (or something)? D. What may (or may not) happen if an event happens (or did not happen)? You can also create other types of question.} \newline \textbf{Context}:I was told, in person over the phone, that my shoes were on their way. They have my money. I have no shoes. \newline \textbf{Question}: \textcolor{red}{What may be the plausible reason for this event?}\\
        \cdashlinelr{1-1}
        \textcolor{brown}{\textit{\textbf{Reframed Task:}}}
        \textit{Use 'what may happen', 'will ...?', 'why might', 'what may have caused', 'what may be true about', 'what is probably true about', 'what must' and similar phrases in your question based on the input context.}\newline \textbf{Context}:I was told, in person over the phone, that my shoes were on their way. They have my money. I have no shoes. \newline \textbf{Question}:\textcolor{blue}{What may happen if my shoes never show up?}\\ [0.1cm]
        \midrule
        \midrule
        \textcolor{brown}{\textit{\textbf{Raw Task:}}} \textit{Craft a question which requires commonsense to be answered. Based on the given context, craft a common-sense question, especially those that are LONG, INTERESTING, and COMPLEX. The goal is to write questions that are easy for humans and hard for AI machines! To create such questions, here are some suggestions: A. What may (or may not) be the plausible reason for an event? B. What may (or may not) happen before (or after, or during) an event? C. What may (or may not) be a plausible fact about someone (or something)? D. What may (or may not) happen if an event happens (or did not happen)? You can also create other types of question.} \newline \textbf{Context}:you see , at my age relationship is kind of important and i thought i got the one after all these years . I noticed that once again i was wrong . i was good simply because i was good , i was caring , helping , supportive , bla bla blaaa . \newline \textbf{Question}: \textcolor{red}{what is the reason why you keep falling for the same type of person?\newline \newline
        A. What is the reason you keep falling for the same type of person?\newline
        B. What may (or may not) be the problem with your relationships?\newline
        C. What may (or may not) be the possible cause}\\
        \cdashlinelr{1-1}
        \textcolor{brown}{\textit{\textbf{Reframed Task:}}}
        \textit{Use 'what may happen', 'will ...?', 'why might', 'what may have caused', 'what may be true about', 'what is probably true about', 'what must' and similar phrases in your question based on the input context.}\newline \textbf{Context}:you see , at my age relationship is kind of important and i thought i got the one after all these years . I noticed that once again i was wrong . i was good simply because i was good , i was caring , helping , supportive , bla bla blaaa . \newline \textbf{Question}:\textcolor{blue}{What may happen if i don't get the one i want ?}\\ [0.1cm]
        \midrule
        \midrule
        \textcolor{brown}{\textit{\textbf{Raw Task:}}} \textit{Craft a question which requires commonsense to be answered. Based on the given context, craft a common-sense question, especially those that are LONG, INTERESTING, and COMPLEX. The goal is to write questions that are easy for humans and hard for AI machines! To create such questions, here are some suggestions: A. What may (or may not) be the plausible reason for an event? B. What may (or may not) happen before (or after, or during) an event? C. What may (or may not) be a plausible fact about someone (or something)? D. What may (or may not) happen if an event happens (or did not happen)? You can also create other types of question.} \newline \textbf{Context}: GPS technology dates back to the time when first ever satellite was launched in the sky in 1979. The era of global positioning started then. \newline \textbf{Question}: \textcolor{red}{What was the first satellite launched in 1979? \newline \newline
        A. Voyager 1 \newline
        B. Voyager 2 \newline
        C. Pioneer 11 \newline
        D. Pioneer 10}\\
        \cdashlinelr{1-1}
        \textcolor{brown}{\textit{\textbf{Reframed Task:}}}
        \textit{Use 'what may happen', 'will ...?', 'why might', 'what may have caused', 'what may be true about', 'what is probably true about', 'what must' and similar phrases in your question based on the input context.}\newline \textbf{Context}:GPS technology dates back to the time when first ever satellite was launched in the sky in 1979. The era of global positioning started then. \newline \textbf{Question}:\textcolor{blue}{Why might GPS technology have been invented??}\\
        \bottomrule
        \end{tabular}
        }
        \caption{
            Examples of {\bf Pattern Reframed} instructions. 
            \textit{Italicized} text represents the instructions. 
            GPT3-instruct's response to the raw and reframed instructions are indicated with \textcolor{red}{red} and \textcolor{blue}{blue} markings, respectively.
        }
    \label{tab:pattern_reframing_both1}
\end{table*}

\clearpage

\begin{table*}[ht]
    \small
    \centering
   %  \resizebox*{!}{0.98\textheight}{%
    \begin{tabular}{p{15cm}}
    \toprule
        \textit{Raw task definitions for tasks requiring \textsc{\textcolor{purple}{Pattern Reframing}}} \\ 
        \midrule
        % \parbox[t]{2mm}{\multirow{9}{*}{\rotatebox[origin=c]{90}{
        %  \textsc{\textcolor{purple}{\parbox{10.5 cm}{\small \centering Pattern \\ Reframing}}}
        %  }}}
        \textcolor{brown}{\textit{\textbf{Raw Task:}}} \textit{Definition: In this task, we ask you to write a question that involves “event duration", based on a given sentence. Here, event duration is defined as the understanding of how long events typically last. For example, “brushing teeth”, usually takes few minutes. \newline Emphasis \& Caution: The written questions are not required to have a single correct answer. \newline Things to avoid: Don't create questions which have explicit mentions of answers in text. Instead, it has to be implied from what is given. In other words, we want you to use "instinct" or "common sense". \newline Prompt: Ask a question on "event duration" based on the provided sentence.} \newline \textbf{Context}: Islam later emerged as the majority religion during the centuries of Ottoman rule, though a significant Christian minority remained. \newline \textbf{Question}: \textcolor{red}{What percentage of Muslims reside in Turkey?}\\
        % \cdashlinelr{2-2}
        % \cdashlinelr{2-2}
        \midrule
        \midrule
        \textcolor{brown}{\textit{\textbf{Raw Task:}}} \textit{Definition: In this task, we ask you to write a question that involves “event duration", based on a given sentence. Here, event duration is defined as the understanding of how long events typically last. For example, “brushing teeth”, usually takes few minutes. \newline Emphasis \& Caution: The written questions are not required to have a single correct answer. \newline Things to avoid: Don't create questions which have explicit mentions of answers in text. Instead, it has to be implied from what is given. In other words, we want you to use "instinct" or "common sense". \newline Prompt: Ask a question on "event duration" based on the provided sentence.} \newline \textbf{Sentence}:About 30\% of Ratners's profit already is derived from the U.S. \newline \textbf{Question}: \textcolor{red}{What are the 3 ways that Ratners derives profits from the U.S.?}\\
        \midrule
        \midrule
        \textcolor{brown}{\textit{\textbf{Raw Task:}}} \textit{Definition: In this task, we ask you to write a question that involves “event duration", based on a given sentence. Here, event duration is defined as the understanding of how long events typically last. For example, “brushing teeth”, usually takes few minutes. \newline Emphasis \& Caution: The written questions are not required to have a single correct answer. \newline Things to avoid: Don't create questions which have explicit mentions of answers in text. Instead, it has to be implied from what is given. In other words, we want you to use "instinct" or "common sense". \newline Prompt: Ask a question on "event duration" based on the provided sentence.} \newline \textbf{Sentence}: Several tenants blame other neighbors as perpetrators of the rift, however. \newline \textbf{Question}: \textcolor{red}{What are the lengths of the events in the sentence?}\\
        \bottomrule
\end{tabular}
% }
\caption{Examples of tasks that require {\bf Pattern Reframing}. 
            \textit{Italicized} text represents the raw instructions. GPT3-instruct's response to the raw instructions is indicated with \textcolor{red}{red} markings.
}
    \label{tab:pattern_reframing_single2}
\end{table*}

\clearpage

%MCTaco Question Generation
\begin{table*}[ht]
    \small
    \centering
   %  \resizebox*{!}{0.98\textheight}{%
    \begin{tabular}{p{15cm}}
    \toprule
        \textit{Raw task definitions, their \textsc{\textcolor{purple}{Pattern Reframed}} counterparts and GPT3-instruct outputs} \\ 
        \midrule
        \textcolor{brown}{\textit{\textbf{Raw Task:}}} \textit{In this task, we ask you to write a question that involves “event duration", based on a given sentence. Here, event duration is defined as the understanding of how long events typically last. For example, “brushing teeth”, usually takes few minutes.} \newline \textbf{Sentence}: Islam later emerged as the majority religion during the centuries of Ottoman rule, though a significant Christian minority remained. \newline \textbf{Question}: \textcolor{red}{What is the event that corresponds to the sentence? \newline
        \newline What is the event that corresponds to the sentence, "Islam later emerged as the majority religion during the centuries of Ottoman rule"?}\\
        \cdashlinelr{1-1}
        \textcolor{brown}{\textit{\textbf{Reframed Task:}}}
        \textit{Use 'how long' or 'what is the duration' or 'for how much time' or similar phrases in your question based on the following context.}\newline \textbf{Sentence}: Islam later emerged as the majority religion during the centuries of Ottoman rule, though a significant Christian minority remained. \newline \textbf{Question}:\textcolor{blue}{What is the duration of Ottoman rule?}\\ [0.1cm]
        \midrule
        \midrule
        \textcolor{brown}{\textit{\textbf{Raw Task:}}} \textit{In this task, we ask you to write a question that involves “event duration", based on a given sentence. Here, event duration is defined as the understanding of how long events typically last. For example, “brushing teeth”, usually takes few minutes.} \newline \textbf{Sentence}:About 30\% of Ratners's profit already is derived from the U.S. \newline \textbf{Question}: \textcolor{red}{How much of Ratner's profit comes from the U.S.?}\\
        \cdashlinelr{1-1}
        \textcolor{brown}{\textit{\textbf{Reframed Task:}}}
        \textit{Use 'how long' or 'what is the duration' or 'for how much time' or similar phrases in your question based on the following context.}\newline \textbf{Sentence}:About 30\% of Ratners's profit already is derived from the U.S. \newline \textbf{Question}:\textcolor{blue}{How long has Ratners been successful in the U.S.?}\\ [0.1cm]
        \midrule
        \midrule
        % \cdashlinelr{2-2}
        % \cdashlinelr{2-2}
        \textcolor{brown}{\textit{\textbf{Raw Task:}}} \textit{In this task, we ask you to write a question that involves “event duration", based on a given sentence. Here, event duration is defined as the understanding of how long events typically last. For example, “brushing teeth”, usually takes few minutes.} \newline \textbf{Sentence}: Several tenants blame other neighbors as perpetrators of the rift, however. \newline \textbf{Question}: \textcolor{red}{What is the average duration of a family feud? }\\
        \cdashlinelr{1-1}
        \textcolor{brown}{\textit{\textbf{Reframed Task:}}}
        \textit{Use 'how long' or 'what is the duration' or 'for how much time' or similar phrases in your question based on the following context.}\newline \textbf{Sentence}: Several tenants blame other neighbors as perpetrators of the rift, however. \newline \textbf{Question}:\textcolor{blue}{How long has this rift been going on?}\\
        \bottomrule
\end{tabular}
% }
\caption{Examples of {\bf Pattern Reframed} instructions. 
            \textit{Italicized} text represents the instructions. 
            GPT3-instruct's response to the raw and reframed instructions are indicated with \textcolor{red}{red} and \textcolor{blue}{blue} markings, respectively.
}
    \label{tab:pattern_reframing_both2}
\end{table*}

\clearpage

\subsubsection{\textsc{Itemizing Reframing}}
Table~\ref{tab:itemizing_reframing_single1} shows how raw instruction in its detailed form can not help GPT3 produce the valid questions for the QASC related fact generation task. Table~\ref{tab:itemizing_reframing_both1} illustrates how reducing the raw instruction content (retaining only the Definition) still does not help model to perform the task and how reframing helps the model to perform the task. Table~\ref{tab:itemizing_reframing_both2} shows how \textsc{Itemizing Reframing} works for some miscellaneous tasks.

%Itemizing reframing
\begin{table*}[ht]
    \small
    \centering
     \resizebox*{!}{0.72\textheight}{
    \begin{tabular}{p{15cm}}
    \toprule
        \textit{Raw task definitions for tasks requiring \textsc{\textcolor{purple}{Itemizing Reframing}}} \\ 
        \midrule
        % \parbox[t]{2mm}{\multirow{9}{*}{\rotatebox[origin=c]{90}{
        %  \textsc{\textcolor{purple}{\parbox{18.5cm}{\small \centering Itemizing \\ Reframing}}}
        %  }}}
        \textcolor{brown}{\textit{\textbf{Raw Task:}}} \textit{Definition: Write a fact related to the given fact. Note that, your fact should have at least one word in common with the given fact. All facts in this task refer to science facts. Your related fact must form a chain with the given fact. Chains form when two facts connect together to produce a third fact. An example of a chain is: "pesticides cause pollution" (given fact) + "pollution can harm animals" (related fact) → "pesticides can harm animals" (connected chain) \newline
        Emphasis \& Caution: It's okay if your related fact includes some irrelevant information, provided that it has some overlap with the given fact and it contains some words corresponding to the provided topic. \newline
        Things to avoid: Avoid creating simple paraphrases of the given fact. While your generated fact should be related to the input fact, they must describe slightly different scientific phenomenon. \newline
        Prompt: Write a related fact to a given fact} \newline \textbf{Fact:} an earthquake causes the ground to shake \newline \textbf{Fact:} \textcolor{red}{The number of earthquakes detected by seismic stations between 2000 and 2009 \newline
        Fact: the frequency of earthquakes worldwide \newline
        Fact: The frequency of earthquakes worldwide \newline
        Fact: 2010 was an active year for earthquakes \newline
        Fact: The number of earthquakes detected by seismic stations between 2000 and 2009 \newline
        Fact: the frequency of earthquakes worldwide \newline
        Fact:}\\
        \midrule
        \midrule
        % \midrule
        \textcolor{brown}{\textit{\textbf{Raw Task:}}} \textit{Definition: Write a fact related to the given fact. Note that, your fact should have at least one word in common with the given fact. All facts in this task refer to science facts. Your related fact must form a chain with the given fact. Chains form when two facts connect together to produce a third fact. An example of a chain is: "pesticides cause pollution" (given fact) + "pollution can harm animals" (related fact) → "pesticides can harm animals" (connected chain) \newline
        Emphasis \& Caution: It's okay if your related fact includes some irrelevant information, provided that it has some overlap with the given fact and it contains some words corresponding to the provided topic. \newline
        Things to avoid: Avoid creating simple paraphrases of the given fact. While your generated fact should be related to the input fact, they must describe slightly different scientific phenomenon. \newline
        Prompt: Write a related fact to a given fact} \newline \textbf{Fact:} a solar panel converts sunlight into electricity. \newline \textbf{Fact:} \textcolor{red}{a light bulb converts electricity into light. \newline
        Related fact: a solar panel converts sunlight into electricity. \newline
        Related fact: a light bulb converts electricity into light. \newline
        Related fact: a solar panel can convert sunlight into electricity. \newline
        Related fact: a light bulb can convert electricity into light. \newline
        What is the difference between}\\
        \midrule
        \midrule
        \textcolor{brown}{\textit{\textbf{Raw Task:}}} \textit{Definition: Write a fact related to the given fact. Note that, your fact should have at least one word in common with the given fact. All facts in this task refer to science facts. Your related fact must form a chain with the given fact. Chains form when two facts connect together to produce a third fact. An example of a chain is: "pesticides cause pollution" (given fact) + "pollution can harm animals" (related fact) → "pesticides can harm animals" (connected chain) \newline
        Emphasis \& Caution: It's okay if your related fact includes some irrelevant information, provided that it has some overlap with the given fact and it contains some words corresponding to the provided topic. \newline
        Things to avoid: Avoid creating simple paraphrases of the given fact. While your generated fact should be related to the input fact, they must describe slightly different scientific phenomenon. \newline
        Prompt: Write a related fact to a given fact} \newline \textbf{Fact:} damming a river can cause a lake to form. \newline \textbf{Fact:} \textcolor{red}{dams can cause pollution  \newline
        Related Fact: producing fossil fuels causes pollution. \newline
        Related Fact: making fossil fuels causes pollution. \newline
        Fact: lakes can be polluted. \newline
        Fact: lakes can be polluted by dams. \newline
        Related Fact: lakes can be polluted by dams. \newline
        Related Fact: lakes can be polluted by dams that}\\
        \bottomrule
\end{tabular}
}
\caption{Examples of tasks that require {\bf Itemizing Reframing}. \textit{Italicized} text represents the raw instructions. GPT3-instruct's response to the raw instructions is indicated with \textcolor{red}{red} markings.
}
    \label{tab:itemizing_reframing_single1}
\end{table*}

\clearpage

\begin{table*}[ht]
    \small
    \centering
   %  \resizebox*{!}{0.98\textheight}{%
    \begin{tabular}{p{15cm}}
    \toprule
        \textit{Raw task definitions, their \textsc{\textcolor{purple}{Itemizing Reframed}} counterparts and GPT3-instruct outputs} \\ 
        \midrule
        % \parbox[t]{2mm}{\multirow{9}{*}{\rotatebox[origin=c]{90}{
        %  \textsc{\textcolor{purple}{\parbox{18 cm}{\small \centering Itemizing \\ Reframing}}}
        %  }}}
        \textcolor{brown}{\textit{\textbf{Raw Task:}}} \textit{Write a fact related to the given fact. Note that, your fact should have at least one word in common with the given fact. All facts in this task refer to science facts. Your related fact must form a chain with the given fact. Chains form when two facts connect together to produce a third fact. An example of a chain is: "pesticides cause pollution" (given fact) + "pollution can harm animals" (related fact) → "pesticides can harm animals" (connected chain)} \newline \textbf{Fact:} an earthquake causes the ground to shake \newline \textbf{Fact}: \textcolor{red}{a tsunami is a large wave in the ocean.}\\
        \cdashlinelr{1-1}
        \textcolor{brown}{\textit{\textbf{Reframed Task:}}}
        \textit{Write a fact related to the given fact. \newline
        - Note that, your fact should have at least one word in common with the given fact. \newline
        - All facts in this task refer to science facts. \newline
        - Your related fact must form a chain with the given fact. Chains form when two facts connect together to produce a third fact. An example of a chain is: "pesticides cause pollution" (given fact) + "pollution can harm animals" (related fact) → "pesticides can harm animals" (connected chain)}\newline \textbf{Fact:} an earthquake causes the ground to shake \newline \textbf{Fact:} \textcolor{blue}{earthquakes may cause damage to buildings.}\\ [0.1cm]
        % \cdashlinelr{2-2}
        % \cdashlinelr{2-2}
        \midrule
        \midrule
        \textcolor{brown}{\textit{\textbf{Raw Task:}}} \textit{Write a fact related to the given fact. Note that, your fact should have at least one word in common with the given fact. All facts in this task refer to science facts. Your related fact must form a chain with the given fact. Chains form when two facts connect together to produce a third fact. An example of a chain is: "pesticides cause pollution" (given fact) + "pollution can harm animals" (related fact) → "pesticides can harm animals" (connected chain)} \newline \textbf{Fact:} a solar panel converts sunlight into electricity. \newline \textbf{Question}: \textcolor{red}{ the sun is the center of our solar system \newline \newline The sun is the center of our solar system.}\\
        \cdashlinelr{1-1}
        \textcolor{brown}{\textit{\textbf{Reframed Task:}}}
        \textit{Write a fact related to the given fact. \newline
        - Note that, your fact should have at least one word in common with the given fact. \newline
        - All facts in this task refer to science facts. \newline
        - Your related fact must form a chain with the given fact. Chains form when two facts connect together to produce a third fact. An example of a chain is: "pesticides cause pollution" (given fact) + "pollution can harm animals" (related fact) → "pesticides can harm animals" (connected chain)}\newline \textbf{Fact:} a solar panel converts sunlight into electricity. \newline \textbf{Fact:} \textcolor{blue}{Solar panels can be used to power homes and businesses.}\\ [0.1cm]
        \midrule
        \midrule
        % \cdashlinelr{2-2}
        % \cdashlinelr{2-2}
        \textcolor{brown}{\textit{\textbf{Raw Task:}}} \textit{Write a fact related to the given fact. Note that, your fact should have at least one word in common with the given fact. All facts in this task refer to science facts. Your related fact must form a chain with the given fact. Chains form when two facts connect together to produce a third fact. An example of a chain is: "pesticides cause pollution" (given fact) + "pollution can harm animals" (related fact) → "pesticides can harm animals" (connected chain)} \newline \textbf{Fact}: damming a river can cause a lake to form. \newline \textbf{Fact}: \textcolor{red}{when the water level is high, fish are at risk \newline
        Fact: when the water level is low, fish are at risk
        Fact: when the water level is high, fish are at risk
        Write a fact related to the given fact. Note that, your fact should have at least one word in common with the}\\
        \cdashlinelr{1-1}
        \textcolor{brown}{\textit{\textbf{Reframed Task:}}}
        \textit{Write a fact related to the given fact. \newline
        - Note that, your fact should have at least one word in common with the given fact. \newline
        - All facts in this task refer to science facts. \newline
        - Your related fact must form a chain with the given fact. Chains form when two facts connect together to produce a third fact. An example of a chain is: "pesticides cause pollution" (given fact) + "pollution can harm animals" (related fact) → "pesticides can harm animals" (connected chain)}\newline \textbf{Fact}: damming a river can cause a lake to form. \newline \textbf{Fact}:\textcolor{blue}{dams can cause algae to grow.}\\
        \bottomrule
\end{tabular}
% }
\caption{Examples of {\bf Itemizing Reframed} instructions. 
            \textit{Italicized} text represents the instructions. 
            GPT3-instruct's response to the raw and reframed instructions are indicated with \textcolor{red}{red} and \textcolor{blue}{blue} markings, respectively.
}
    \label{tab:itemizing_reframing_both1}
\end{table*}

\clearpage

\begin{table*}[ht]
    \small
    \centering
   %  \resizebox*{!}{0.98\textheight}{%
    \begin{tabular}{p{15cm}}
    \toprule
         \textit{Raw task definitions, their \textsc{\textcolor{purple}{Itemizing Reframed}} counterparts and GPT3-instruct outputs} \\ 
        \midrule
        % \parbox[t]{2mm}{\multirow{9}{*}{\rotatebox[origin=c]{90}{
        %  \textsc{\textcolor{purple}{\parbox{22.5 cm}{\small \centering Itemizing \\ Reframing}}}
        %  }}}
        \textcolor{brown}{\textit{\textbf{Raw Task:}}} \textit{Write a story. Mention John in the story.  Mention Hari} \newline \textbf{Story}: \textcolor{red}{There was an old woman who lived in the woods. She was very lonely and had not seen another human for many years. She was very excited because she had found an abandoned house, and she had found companionship with a dog that had come to live with her. She had cleaned up the old house, and}\\
        \cdashlinelr{1-1}
        \textcolor{brown}{\textit{\textbf{Reframed Task:}}}
        \textit{Write a story. \newline
        -Mention John in the story. \newline 
        -Mention Hari.} \newline \textbf{Story} \textcolor{blue}{John and Hari went to the store together. John was going to buy a cake for his family, and Hari was going to buy a new shirt. They got into the store and never saw each other again.}\\ [0.1cm]
        \midrule
        \midrule
        % \midrule
        \textcolor{brown}{\textit{\textbf{Raw Task:}}} \textit{Write a sentence. It should start with 'What'. It should not have more than 4 words. It should mention "Patrick".} \newline \textbf{Sentence}: \textcolor{red}{ What is Patrick's favorite color?}\\
        \cdashlinelr{1-1}
        \textcolor{brown}{\textit{\textbf{Reframed Task:}}}
        \textit{Write a sentence. \newline
        - It should start with 'What'. \newline
        - It should have a maximum of 4 words. \newline
        - It should mention "Patrick".} \newline \textbf{Sentence}: \textcolor{blue}{What does Patrick do?}\\ [0.1cm]
        \midrule
        \midrule
        \textcolor{brown}{\textit{\textbf{Raw Task:}}} \textit{Write a story based on the following passage. The story should not be longer than the passage. Don't use the names mentioned in the passage}\newline \textbf{Passage}: The earthquake swarm was noted on October 12, 2007 in the Prince George Citizen by citizen staff, three days after the earthquakes began. Scientists mentioned in the report were seismologist John Cassidy of Natural Resources Canada and volcanologist Catherine Hickson, who was part of the Geological Survey of Canada at the time. At the time of the report, scientists did not know the origin of the swarm. Seismologist John Cassidy stated, "the depth is enough to rule out hydrothermal but it's up in the air as to whether the cause is tectonic shifts or volcanic activity. If it is volcanic there are certain characteristics that we would expect, there's a tremor-like character to it. And so we'll be looking for the types of events that we see beneath volcanoes and we'll be looking to see if they're getting closer to the surface or if they're migrating at all."Even if the Nazko swarm were a warning of a volcanic eruption, Hickson doubted it would turn out to be a highly explosive eruption like those that can occur in subduction-zone volcanoes. "We're not talking about an injection of tonnes of ash many kilometers into the air like the 1980 Mount St. Helens eruption or the 1991 Mount Pinatubo eruption. We're talking about something very small, relatively localized that should have a fairly limited impact... but it'll be extremely exciting", Hickson said. If an eruption were to occur, Hickson suggested that it would be characterized by a lava fountain that sends globs of lava 100 m (330 ft) into the air. This is similar to those that occur in Hawaii. Hickson said that a Nazko eruption could be a tourist attraction, but warned that noxious gases such as carbon dioxide and sulfur dioxide would be released during the event \newline \textbf{Story}: \textcolor{red}{John Cassidy, a seismologist, and Catherine Hickson, a volcanologist, are at the office of Natural Resources Canada when John receives a call. John has to take the call, so Catherine goes to the breakroom to get herself some coffee. While she is waiting for the coffee to brew, she hears a faint sound coming from the kitchen. She walks over to investigate and sees the coffee pot shaking on the counter. She goes to turn it off just as the phone rings.}\\
        \cdashlinelr{1-1}
        \textcolor{brown}{\textit{\textbf{Reframed Task:}}}
        \textit{Write a story based on the following passage. \newline
        - The story should be shorter than the passage. \newline
        - Use new names}\newline \textbf{Passage}: The earthquake swarm was noted on October 12, 2007 in the Prince George Citizen by citizen staff, three days after the earthquakes began. Scientists mentioned in the report were seismologist John Cassidy of Natural Resources Canada and volcanologist Catherine Hickson, who was part of the Geological Survey of Canada at the time. At the time of the report, scientists did not know the origin of the swarm. Seismologist John Cassidy stated, "the depth is enough to rule out hydrothermal but it's up in the air as to whether the cause is tectonic shifts or volcanic activity. If it is volcanic there are certain characteristics that we would expect, there's a tremor-like character to it. And so we'll be looking for the types of events that we see beneath volcanoes and we'll be looking to see if they're getting closer to the surface or if they're migrating at all."Even if the Nazko swarm were a warning of a volcanic eruption, Hickson doubted it would turn out to be a highly explosive eruption like those that can occur in subduction-zone volcanoes. "We're not talking about an injection of tonnes of ash many kilometers into the air like the 1980 Mount St. Helens eruption or the 1991 Mount Pinatubo eruption. We're talking about something very small, relatively localized that should have a fairly limited impact... but it'll be extremely exciting", Hickson said. If an eruption were to occur, Hickson suggested that it would be characterized by a lava fountain that sends globs of lava 100 m (330 ft) into the air. This is similar to those that occur in Hawaii. Hickson said that a Nazko eruption could be a tourist attraction, but warned that noxious gases such as carbon dioxide and sulfur dioxide would be released during the event \newline \textbf{Story}:\textcolor{blue}{A year ago, in October 2007, an earthquake swarm began in the Nazko region of Canada. Scientists were unsure of the origins of the swarm, but they were certain it couldn't be volcanic. There were three possible origins, all of which were ruled out by the seismologists. The swarm was getting closer to the surface however, and it was a mystery as to why. Scientists were closely monitoring the swarm for any changes or any news of a possible eruption.}\\
        \bottomrule
\end{tabular}
% }
\caption{Examples of {\bf Itemizing Reframed} instructions. 
            \textit{Italicized} text represents the instructions. 
            GPT3-instruct's response to the raw and reframed instructions are indicated with \textcolor{red}{red} and \textcolor{blue}{blue} markings, respectively.
}
    \label{tab:itemizing_reframing_both2}
\end{table*}

\clearpage

\subsubsection{\textsc{Decomposition Reframing}}
Table~\ref{tab:decomposition_reframing_single1} shows how raw instruction in its detailed form as well as with reduced form(definition only) can not help GPT3 produce the valid questions for the Winogrande sample generation task. Table~\ref{tab:decomposition_reframing_both1} illustrates how reframing helps the model to perform the task.

\begin{table*}[ht]
    \small
    \centering
     \resizebox*{!}{0.73\textheight}{%
    \begin{tabular}{p{15cm}}
    \toprule
        \textit{Raw task definitions for tasks requiring \textsc{\textcolor{purple}{Decomposition Reframing}}} \\ 
        \midrule
        % \parbox[t]{2mm}{\multirow{9}{*}{\rotatebox[origin=c]{90}{
        %  \textsc{\textcolor{purple}{\parbox{19.5cm}{\small \centering Decomposition \\ Reframing}}}
        %  }}}
        \textcolor{brown}{\textit{\textbf{Raw Task:}}} \textit{In this task, based on the given context word, you need to create a pair of sentences each containing a blank (\_) and their corresponding answer. The sentence pair should look similar, and should be about two related but different objects; for example "trophy" and "suitcase". Also, the sentences must be different in terms of trigger words (e.g., "small" and "big") which express contrasting properties about the two objects. }\newline
        \textbf{Context word}:fitness
        \newline \textcolor{red} {
        -The machine's objective is to \newline
        -stimulate \newline
        -the brain cells. \newline
        -It does this by  \newline
        -stimulating \newline
        -the cells. \newline
        -The machine's objective is to \newline
        -stimulate \newline
        -the}\\
        \midrule
        \midrule
        \textcolor{brown}{\textit{\textbf{Raw Task:}}} \textit{Definition: In this task, based on the given context word, you are asked to create a pair of sentences each containing a blank (\_) and their corresponding answer. The sentence pair should look similar, and should be about two related but different objects; for example "trophy" and "suitcase". Additionally, the two sentences must be different in terms of trigger words (e.g., "small" and "big") which express contrasting properties about the two objects. \newline
        Emphasis \& Caution: 1. Both twin sentences must contain at least 15 and at most 30 words. 2. Twin sentences must have at least 70\% overlapping words. 3. You must utilize the given context word while writing the twin sentences. 4. Each of the twin sentences must contain only one blank. 5. Make sure that ObjectX and Y have the same number e.g. when ObjectX is singular, ObjectY must be singular, too. 6. The two objects (ObjectX \& ObjectY) should be used ONCE in each sentence. 7. Here is a list of contrastive words that may be used as trigger words. You should create more such trigger words and use them in your twin sentences. | Attribute | triggerword | contrastive triggerword | | age | old | new | | altitude | low | high | | area | small | vast | | brightness | dark | light | | clarity | obscure | clear | | cleanness | dirty | clean | | complexity | simple | complex | | cost | cheap | expensive | | density | sparse | dense | | depth | shallow | deep | | distance | near | far | | electric conductivity | low | high | | flexibility | rigid | flexible | | granularity | fine | coarse | | hardness | soft | hard | | length | short | long | | magnitude | small | large | | mass | small | large | | odor | weak | strong | | pressure | low | high | | resistance | low | high | | shape | round | sharp | | shape | flat | spiky | | size | small | large | | sound | quiet | loud | | sound pitch | low | high | | speed | slow | fast | | stability | unstable | stable | | strength | weak | strong | | temperature | low | high | | texture | smooth | rough | | thermal conductivity | low | high | | thickness | thin | thick | | volume | small | large | | weight | light | heavy | | width | narrow | wide | | location | in | out | | location | up | down | | location | above | below | | location | on | off | | location | to | from | \newline
        Things to avoid: 1. The answer must not be associated with the trigger word; instead it should depend on the context present in twin sentences. 2. The answers should not be ambiguous. For each sentence, there should be an agreed upon answer to fill in the blank. 3. Your generations should NOT contain potentially explicit, offensive, or adult content. 4. Do not use animals as your objects. 5. Your objects should not be proper nouns (e.g., New York, Macbook, Jeff Bezos, McDonald's, ...). 6. Avoid repeating the same style or phrase in twin sentences e.g., a contrasting sentence can always be created using simple negation i.e. by adding not, never, etc. Instead, try to increase diversity. \newline
        Prompt: Create a pair of fill in the blanks question-answer pairs based on the context word.)} \newline \textbf{Context Word}: fitness. \newline
        \textcolor{red}{1. It was a common belief that one should not partake in any physical activity \newline
        between meal times, because it would impede digestion. \newline
        \_ \newline
        2. Exercise was recommended because it would aid digestion. \newline
        \_  \newline
        3. Exercise was not recommended because it would inhibit digestion. \newline \newline
        1. It was a common belief that one should not partake in any physical activity \newline
        between meal times, because it would impede digestion. \newline
        e.g., \newline
        2. Exercise was recommended because it would aid digestion. \newline
        e.g., \newline
        3. Exercise was not recommended because it would inhibit}\\
        \bottomrule
\end{tabular}
}
\caption{Examples of tasks that require {\bf Decomposition Reframing}. \textit{Italicized} text represents the raw instructions. GPT3-instruct's response to the raw instructions is indicated with \textcolor{red}{red} markings.
}
    \label{tab:decomposition_reframing_single1}
\end{table*}

\clearpage

\begin{table*}[ht]
    \small
    \centering
   %  \resizebox*{!}{0.98\textheight}{%
    \begin{tabular}{p{15cm}}
    \toprule
         \textit{Raw task definitions, their \textsc{\textcolor{purple}{Decomposition Reframed}} counterparts and GPT3-instruct output} \\ 
        \midrule
        % \parbox[t]{2mm}{\multirow{9}{*}{\rotatebox[origin=c]{90}{
        %  \textsc{\textcolor{purple}{\parbox{21.5cm}{\small \centering Decomposition \\ Reframing}}}
        %  }}}
        \textcolor{brown}{\textit{\textbf{Raw Task:}}} \textit{In this task, based on the given context word, you need to create a pair of sentences each containing a blank (\_) and their corresponding answer. The sentence pair should look similar, and should be about two related but different objects; for example "trophy" and "suitcase". Also, the sentences must be different in terms of trigger words (e.g., "small" and "big") which express contrasting properties about the two objects. \newline
         \textbf{Context word}:fitness
         \newline \textcolor{red} {He realized that he needed to \newline
         He realized that he needed to \newline \newline
         find a new workout routine.\newline
         find a new workout routine. \newline \newline
         I couldn't believe how \newline
         I couldn't believe how \newline \newline
         good he looked. \newline
         good he looked.} }\\
        \cdashlinelr{1-1}
        \textcolor{brown}{\textit{\textbf{Reframed Task 1:}}}
        \textit{Write 2 objects based on the given context word.} \newline \textbf{Context word:} fitness \newline
        \textbf{Answer}: \textcolor{blue}{Fitness tracker \newline Pilates mat} \newline
        \textcolor{brown}{\textit{\textbf{Reframed Task 2:}}}
        \textit{Write a sentence by connecting both objects with a verb.} \newline \textbf{Objects:} Fitness tracker, Pilates mat \newline
        \textcolor{blue}{I've been using my fitness tracker to monitor my progress with Pilates mat.} \newline
        \textcolor{brown}{\textit{\textbf{Reframed Task 3:}}}
        \textit{Create a fill in the blank question from the sentence where object 1 will fill in the blank} \newline \textbf{Object 1}: Trophy \newline
        \textbf{Sentence}: The trophy does not fit the brown suitcase. \newline
        \textbf{Question}: The trophy does not fit the brown suitcase because \_ is large. \newline
        \textbf{Answer}: Trophy \newline
        <other similar examples> \newline
        \textbf{Object 1}: Fitness tracker. \newline
        \textbf{Sentence}: I've been using my fitness tracker to monitor my progress with Pilates mat \newline
        \textbf{Question}: \textcolor{blue} {I've been using my fitness tracker to monitor my progress with Pilates mat because \_ is a good way to keep track. \textbf{Answer:} Fitness tracker.} \newline
        \textcolor{brown}{\textit{\textbf{Reframed Task 4:}}}
        \textit{Change the question such that answer flips to object 2 in the question} \newline \textbf{Object 2}: Suitcase \newline
        \textbf{Sentence}: The trophy does not fit the brown suitcase. \newline
        \textbf{Question}: The trophy does not fit the brown suitcase because \_ is large. \newline \textbf{Answer}: Trophy \newline
        \textbf{Modified Question}:  The trophy does not fit the brown suitcase because \_ is small. \newline \textbf{Answer}: Suitcase \newline 
        <other similar examples> \newline
        \textbf{Object 2}: Pilates mat \newline
        \textbf{Sentence}: I've been using my fitness tracker to monitor my progress with Pilates mat \newline
        \textbf{Question}:  I've been using my fitness tracker to monitor my progress with Pilates mat because \_ is a good way to keep track. \newline
        \textbf{Answer}: Fitness tracker  \newline
        \textbf{Modified Question}:  \textcolor{blue} {I've been using my fitness tracker to monitor my progress with Pilates mat because \_ is a good workout. \newline \textbf{Answer}: Pilates mat} \newline
        \textcolor{brown}{\textit{\textbf{Reframed Task 5:}}}
        \textit{Generate both questions and answers} \newline 
        \textbf{Question 1}: The trophy does not fit the brown suitcase because \_ is large. \newline \textbf{Answer 1}: Trophy \newline
        \textbf{Question 2}:  The trophy does not fit the brown suitcase because \_ is small. \newline \textbf{Answer 2}: Suitcase \newline
        <other similar examples> \newline
        \textbf{Question 1}: \textcolor{blue} {I've been using my fitness tracker to monitor my progress with Pilates mat because \_ is a good way to keep track. \newline
        \textbf{Answer 1}: Fitness tracker \newline
        \textbf{Question 2}:  \textcolor{blue} {I've been using my fitness tracker to monitor my progress with Pilates mat because \_ is a good workout. \newline \textbf{Answer 2}: Pilates mat}}\\
        \bottomrule
\end{tabular}
% }
\caption{Examples of {\bf Decomposition Reframed} instructions. 
            \textit{Italicized} text represents the instructions. 
            GPT3-instruct's response to the raw and reframed instructions are indicated with \textcolor{red}{red} and \textcolor{blue}{blue} markings, respectively.
}
    \label{tab:decomposition_reframing_both1}
\end{table*}

\clearpage

\subsubsection{\textsc{Restraining Reframing}}
Table~\ref{tab:restraining_reframing_both1} illustrates how raw instruction can not help GPT3 produce the valid answers for the DROP answer type generation task and how reframing helps GPT3 to perform the task. Table \ref{tab:restraining_reframing_both2} illustrates the utility of \textsc{Restraining Reframing} for various tasks of diverse types.

% restraining reframing
\begin{table*}[ht]
    \small
    \centering
     \resizebox*{!}{0.73\textheight}{%
    \begin{tabular}{p{17.5cm}}
    \toprule
         \textit{Raw task definitions, their \textsc{\textcolor{purple}{Restraining Reframed}} counterparts and GPT3-instruct output} \\ 
        \midrule
        % \parbox[t]{2mm}{\multirow{9}{*}{\rotatebox[origin=c]{90}{
        %  \textsc{\textcolor{purple}{\parbox{22 cm}{\small \centering Restraining \\ Reframing}}}
        %  }}}
        \textcolor{brown}{\textit{\textbf{Raw Task:}}} \textit{This task involves annotating the answer type to a given question that involve some kind of complex reasoning (including numerical reasoning). Note that the questions require looking at more than one part of the passage to answer. There are 3 possible answer types (i) spans, (ii) numbers and (iii) dates. If the answer can be found in the passage, label it as "span". If the answer is a number, label as "number". Similarly, label "date" if you think the answer to the given question is a date.} \newline \textbf{Passage:} The outbreak of the Seven Years' War in Europe in 1756 resulted in renewed conflict between French and British forces in India. The Third Carnatic War spread beyond southern India and into Bengal where British forces captured the French settlement of Chandernagore in 1757. However, the war was decided in the south, where the British successfully defended Madras, and Sir Eyre Coote decisively defeated the French, commanded by Comte de Lally at the Battle of Wandiwash in 1760. After Wandiwash, the French capital of Pondicherry fell to the British in 1761. The war concluded with the signing of the Treaty of aris in 1763, which returned Chandernagore and Pondichéry to France, and allowed the French to have "factories" in India but forbade French traders from administering them. The French agreed to support British client governments, thus ending French ambitions of an Indian empire and making the British the dominant foreign power in India. \newline \textbf{Question:} How many years did the British fight with the French before signing the Treaty of Paris? \newline \textbf{Answer}: \textcolor{red}{The British fought with the French for at least 7 years before signing the Treaty of Paris in 1763.}\\
        \cdashlinelr{1-1}
        \textcolor{brown}{\textit{\textbf{Reframed Task:}}}
        \textit{This task involves annotating the answer type to a given question that involve some kind of complex reasoning (including numerical reasoning). Note that the questions require looking at more than one part of the passage to answer. There are 3 possible answer types (i) spans, (ii) numbers and (iii) dates. If the answer can be found in the passage, label it as "span". If the answer is a number, label as "number". Similarly, label "date" if you think the answer to the given question is a date.} \newline \textbf{Passage:} The outbreak of the Seven Years' War in Europe in 1756 resulted in renewed conflict between French and British forces in India. The Third Carnatic War spread beyond southern India and into Bengal where British forces captured the French settlement of Chandernagore in 1757. However, the war was decided in the south, where the British successfully defended Madras, and Sir Eyre Coote decisively defeated the French, commanded by Comte de Lally at the Battle of Wandiwash in 1760. After Wandiwash, the French capital of Pondicherry fell to the British in 1761. The war concluded with the signing of the Treaty of aris in 1763, which returned Chandernagore and Pondichéry to France, and allowed the French to have "factories" in India but forbade French traders from administering them. The French agreed to support British client governments, thus ending French ambitions of an Indian empire and making the British the dominant foreign power in India. \newline \textbf{Question:} How many years did the British fight with the French before signing the Treaty of Paris? \newline
        Answer either Number, Date or Span based on the answer type. \newline
        \textbf{Answer}: \textcolor{blue}{Number.}\\ [0.1cm]
        \midrule
        \midrule
        \textcolor{brown}{\textit{\textbf{Raw Task:}}} \textit{This task involves annotating the answer type to a given question that involve some kind of complex reasoning (including numerical reasoning). Note that the questions require looking at more than one part of the passage to answer. There are 3 possible answer types (i) spans, (ii) numbers and (iii) dates. If the answer can be found in the passage, label it as "span". If the answer is a number, label as "number". Similarly, label "date" if you think the answer to the given question is a date.} \newline \textbf{Passage:} From 1975, Flavin installed permanent works in Europe and the United States, including "Untitled".In memory of Urs Graf" at the Kunstmuseum Basel (conceived 1972, realized 1975) ; the Kröller-Müller Museum, Otterlo, Netherlands (1977); Hudson River Museum, Yonkers, New York (1979); United States Courthouse, Anchorage, Alaska (1979-89); the Staatliche Kunsthalle Baden-Baden, Germany (1989); the lobby of the MetroTech Center (with Skidmore, Owings \& Merrill), Brooklyn, New York (1992); seven lampposts outside the Städtische Galerie im Lenbachhaus, Munich (1994); Hypovereinsbank, Munich (1995); Institut Arbeit und Technik/Wissenschaftspark, Gelsenkirchen, Germany (1996); and the Union Bank of Switzerland, Bern (1996) . Additional sites for Flavins architectural "interventions" became the Grand Central Station in New York (1976) , Hamburger Bahnhof in Berlin (1996), and the Chinati Foundation in Marfa, Texas (2000). His large-scale work in colored fluorescent light for six buildings at the Chinati Foundation was initiated in the early 1980s, although the final plans were not completed until 1996. His last artwork was a site-specific art at Santa Maria Annunciata in Chiesa Rossa, Milan, Italy. The 1930s church was designed by Giovanni Muzio. The design for the piece was completed two days before Flavins death on November 29, 1996. Its installation was completed one year later with the assistance of the Dia Art Foundation and Fondazione Prada. \newline \textbf{Question:} When did the installation of Santa Maria Annunciata complete? \newline \textbf{Answer}: \textcolor{red}{ 1996}\\
        \cdashlinelr{1-1}
        \textcolor{brown}{\textit{\textbf{Reframed Task:}}}
        \textit{This task involves annotating the answer type to a given question that involve some kind of complex reasoning (including numerical reasoning). Note that the questions require looking at more than one part of the passage to answer. There are 3 possible answer types (i) spans, (ii) numbers and (iii) dates. If the answer can be found in the passage, label it as "span". If the answer is a number, label as "number". Similarly, label "date" if you think the answer to the given question is a date.} \newline \textbf{Passage:} <same as the previous example> \newline \textbf{Question:} When did the installation of Santa Maria Annunciata complete? \newline
        Answer either Number, Date or Span based on the answer type. \newline \textbf{Answer}: \textcolor{blue}{Span}\\ [0.1cm]
        \bottomrule
\end{tabular}
}
\caption{Examples of {\bf Restraining Reframed} instructions. 
            \textit{Italicized} text represents the instructions. 
            GPT3-instruct's response to the raw and reframed instructions are indicated with \textcolor{red}{red} and \textcolor{blue}{blue} markings, respectively.
}
    \label{tab:restraining_reframing_both1}
\end{table*}

\clearpage

\begin{table*}[ht]
    \small
    \centering
   %  \resizebox*{!}{0.98\textheight}{%
    \begin{tabular}{p{15cm}}
    \toprule
        \textit{Raw task definitions for tasks requiring \textsc{\textcolor{purple}{Restraining Reframing}}} \\ 
        \midrule
        % \parbox[t]{2mm}{\multirow{9}{*}{\rotatebox[origin=c]{90}{
        %  \textsc{\textcolor{purple}{\parbox{15.5cm}{\small \centering Restraining \\ Reframing}}}
        %  }}}
        \textcolor{brown}{\textit{\textbf{Raw Task:}}} \textit{Answer an incorrect answer to the given question \newline \textbf{Question:} What is the captial of US?}  \newline \textbf{Answer}: \textcolor{red}{Washington \newline Washington is the capital of the United States of America.}\\
        \cdashlinelr{1-1}
        \textcolor{brown}{\textit{\textbf{Reframed Task:}}}
        \textit{Answer an incorrect answer to the given question }\newline \textbf{Question:} What is the captial of US? \newline
        Answer a city that is not capital of US. \newline
        \textbf{Answer}: \textcolor{blue}{New York}\\ [0.1cm]
        \midrule
        \midrule
        % \midrule
        \textcolor{brown}{\textit{\textbf{Raw Task:}}} \textit{Generate the language of the following text.} \newline \textbf{Question:} What is the captial of US? \newline \textbf{Answer}: \textcolor{red}{ Answer: Washington, D.C. \newline
        What is the capital of the US? \newline
        Washington, D.C.}\\
        \cdashlinelr{1-1}
        \textcolor{brown}{\textit{\textbf{Reframed Task:}}}
        \textit{Generate the language of the following text.} \newline \textbf{Question:} What is the captial of US? \newline Is the language English, Spanish, Geman or any other language? \newline \textbf{Question:} When did the installation of Santa Maria Annunciata complete? \newline \textbf{Answer}: \textcolor{blue}{English}\\ [0.1cm]
        \midrule
        \midrule
        \textcolor{brown}{\textit{\textbf{Raw Task:}}} \textit{Generate the sentence number in the paragraph that contains answer to the given question} \newline \textbf{Paragraph}- Sent 1: A Republican bid to keep the government open past Friday includes no new money for the construction of a border wall along the U.S.-Mexico border, according to several congressional aides familiar with ongoing talks. Sent 2: GOP leaders submitted the new offer Tuesday afternoon in an effort to appease Democrats, whose votes are needed to avert a shutdown of federal agencies, several House and Senate aides said. Sent 3: However, Republicans also insisted on increases in border security and defense spending, including money to repair existing fencing and new surveillance technology to patrol the nearly 2,000-mile border. Sent 4: Democrats have indicated that they would support such appropriations so long as no money goes toward an actual wall. \newline \textbf{Question}: Who has to be appeased to keep the government open? \newline \textbf{Answer}: \textcolor{red}{Democrats have indicated that they would support such appropriations so long as no money goes toward an actual wall.}\\
        \cdashlinelr{1-1}
        \textcolor{brown}{\textit{\textbf{Reframed Task:}}}
        \textit{} \newline \textbf{Paragraph}- Sent 1: A Republican bid to keep the government open past Friday includes no new money for the construction of a border wall along the U.S.-Mexico border, according to several congressional aides familiar with ongoing talks. Sent 2: GOP leaders submitted the new offer Tuesday afternoon in an effort to appease Democrats, whose votes are needed to avert a shutdown of federal agencies, several House and Senate aides said. Sent 3: However, Republicans also insisted on increases in border security and defense spending, including money to repair existing fencing and new surveillance technology to patrol the nearly 2,000-mile border. Sent 4: Democrats have indicated that they would support such appropriations so long as no money goes toward an actual wall. \newline \textbf{Question}: Who has to be appeased to keep the government open? Generate the sentence number in the paragraph that contains answer to the given question\newline Generate the sentence number in the paragraph that contains answer to the given question \newline \textbf{Answer}: \textcolor{blue}{Sent 2}\\
        \bottomrule
\end{tabular}
% }
\caption{Examples of {\bf Restraining Reframed} instructions. 
        \textit{Italicized} text represents the instructions. 
        GPT3-instruct's response to the raw and reframed instructions are indicated with \textcolor{red}{red} and \textcolor{blue}{blue} markings, respectively. 
}
    \label{tab:restraining_reframing_both2}
\end{table*}

\clearpage

\subsubsection{\textsc{Specialization Reframing}}
For a diverse set of tasks, Table~\ref{tab:specialization_reframing_both1} shows how GPT3 fails to perform the task with raw instructions and how \textsc{Specialization Reframing} helps to overcome those failures.

% specialization reframing
\begin{table*}[ht]
    \small
    \centering
   %  \resizebox*{!}{0.98\textheight}{%
    \begin{tabular}{p{15cm}}
    \toprule
         \textit{Raw task definitions, their \textsc{\textcolor{purple}{Specialization Reframed}} counterparts and GPT3-instruct output} \\ 
        \midrule
        % \parbox[t]{2mm}{\multirow{9}{*}{\rotatebox[origin=c]{90}{
        %  \textsc{\textcolor{purple}{\parbox{20.5cm}{\small \centering Pattern \\ Reframing}}}
        %  }}}
        \textcolor{brown}{\textit{\textbf{Raw Task:}}} \textit{Answer the following question.} \newline
        Father's age is 70 years. What could be the son's age? (i) 71 (ii) 72 (iii) 40 (iv) 102? \newline Note that a son's age is always lower than father's age.
        \newline
        \textbf{Answer}: \textcolor{red}{In the given problem the father's age is 70 years. So, the son's age can be either less than 70 years or 70 years. Since the given age is less than 70 years, the son's age is 71 years.} 
         \\ 
        \cdashlinelr{1-1}
        \textit{Select the correct answer among the options provided for the following question. The answer to this type of question will always be a number either smaller or larger than the number mentioned in the question.} \newline
        Father's age is 70 years. What could be the son's age? (i) 71 (ii) 72 (iii) 40 (iv) 102? \newline Note that a son's age is always lower than father's age.
        \newline
        \textbf{Answer}: \textcolor{blue}{(iii) 40} 
         \\ [0.1cm]
        \midrule
        \midrule
        % \rule{0pt}{0.3cm} 
        \textcolor{brown}{\textit{\textbf{Raw Task:}}} \textit{Answer the following question.} \newline \textbf{Context}: Birds are good at flying. However, in the movie, none of the birds could fly. Surprisingly, tigers could fly. \newline \textbf{Question}:  In the movie, who can fly (i) birds or (ii) tigers  \newline \textbf{Answer}: \textcolor{red}{(i) Birds}\\
        \cdashlinelr{1-1}
        \textcolor{brown}{\textit{\textbf{Reframed Task:}}} \textit{Answer the following question only based on the information provided in the context.} \newline \textbf{Context}: Birds are good at flying. However, in the movie, none of the birds could fly. Surprisingly, tigers could fly. \newline \textbf{Question}:  In the movie, who can fly (i) birds or (ii) tigers  \newline \textbf{Answer}: \textcolor{blue}{None of the birds could fly. Surprisingly, tigers could fly.}\\ [0.1cm]
        \midrule
        \midrule
        \textcolor{brown}{\textit{\textbf{Raw Task:}}} \textit{Answer the following question.} \newline \textbf{Passage}: Hoping to rebound from their loss to the Patriots, the Raiders stayed at home for a Week 16 duel with the Houston Texans. Oakland would get the early lead in the first quarter as quarterback JaMarcus Russell completed a 20-yard touchdown pass to rookie wide receiver Chaz Schilens. The Texans would respond with fullback Vonta Leach getting a 1-yard touchdown run, yet the Raiders would answer with kicker Sebastian Janikowski getting a 33-yard and a 30-yard field goal. Houston would tie the game in the second quarter with kicker Kris Brown getting a 53-yard and a 24-yard field goal. Oakland would take the lead in the third quarter with wide receiver Johnnie Lee Higgins catching a 29-yard touchdown pass from Russell, followed up by an 80-yard punt return for a touchdown. The Texans tried to rally in the fourth quarter as Brown nailed a 40-yard field goal, yet the Raiders' defense would shut down any possible attempt. 
        \newline \textbf{Answer}: \textcolor{red}{The Texans would tie the game in the second quarter with kicker Kris Brown getting a 53-yard and a 24-yard field goal.}\\
        \cdashlinelr{1-1}
        \textcolor{brown}{\textit{\textbf{Reframed Task:}}} \textit{Calculate the answer to the given question. You need to perform addition, subtraction or counting operation.} \newline \textbf{Passage}: Hoping to rebound from their loss to the Patriots, the Raiders stayed at home for a Week 16 duel with the Houston Texans. Oakland would get the early lead in the first quarter as quarterback JaMarcus Russell completed a 20-yard touchdown pass to rookie wide receiver Chaz Schilens. The Texans would respond with fullback Vonta Leach getting a 1-yard touchdown run, yet the Raiders would answer with kicker Sebastian Janikowski getting a 33-yard and a 30-yard field goal. Houston would tie the game in the second quarter with kicker Kris Brown getting a 53-yard and a 24-yard field goal. Oakland would take the lead in the third quarter with wide receiver Johnnie Lee Higgins catching a 29-yard touchdown pass from Russell, followed up by an 80-yard punt return for a touchdown. The Texans tried to rally in the fourth quarter as Brown nailed a 40-yard field goal, yet the Raiders' defense would shut down any possible attempt. 
        \newline \textbf{Answer}: \textcolor{blue}{4}\\
        \bottomrule
\end{tabular}
% }
\caption{Examples of {\bf Specialization Reframed} instructions. 
        \textit{Italicized} text represents the instructions. 
        GPT3-instruct's response to the raw and reframed instructions are indicated with \textcolor{red}{red} and \textcolor{blue}{blue} markings, respectively. 
}
    \label{tab:specialization_reframing_both1}
\end{table*}

\end{document}